\documentclass[11pt]{article}

\usepackage[preprint]{acl}

\usepackage{times}
\usepackage{latexsym}
\usepackage{soul}

\usepackage[T1]{fontenc}

\usepackage[utf8]{inputenc}

\usepackage{microtype}
\usepackage{comment}

\usepackage{inconsolata}

\usepackage{graphicx}
\graphicspath{{}}
\usepackage{float}

\usepackage{tcolorbox}

\usepackage{xurl}

\usepackage[table,xcdraw]{xcolor}
\usepackage{booktabs}
\usepackage{tabularx}
\usepackage{longtable}
\usepackage{array}
\usepackage{enumitem}

\newcommand{\corpusname}{IndoSoSci}

\title{Injecting Knowledge from Social Science Journals to Improve Indonesian Cultural Understanding by LLMs}

\author{
 \textbf{Adimulya Kartiyasa},  
 \textbf{Bao Gia Cao},  
 \textbf{Boyang Li}  
\\
 Nanyang Technological University, Singapore\\
 \small{
 adimulya.kartiyasa@ntu.edu.sg
 }
}

\begin{document}
\maketitle
\begin{abstract}

Recently there have been intensifying efforts to improve the understanding of Indonesian cultures by large language models (LLMs). An attractive source of cultural knowledge that has been largely overlooked is local journals of social science, which likely contain substantial cultural studies from a native perspective. We present a novel text dataset of journal article passages, created from 151 open-source Indonesian social science journals, called \corpusname. We  demonstrate an effective recipe for injecting Indonesian cultural knowledge therein into LLMs: extracting the facts related to Indonesian culture, and apply retrieval-augmented generation (RAG) with LLM-generated hypothetical documents as queries during retrieval. The proposed recipe yields strong performance gains over several strong baselines on the IndoCulture benchmark. Additionally, by combining \corpusname{} with Indonesian Wikipedia, we set a new state-of-the-art accuracy on the IndoCulture benchmark. Dataset and code will be made available at a later date.
\end{abstract}

\section{Introduction}

Most large language models today are trained with text in predominantly Western languages, which may have created a Western bias in these models  \citep{caoAssessingCrossCulturalAlignment2023, adilazuardaMeasuringModelingCulture2024, loveniaSEACrowdMultilingualMultimodal2024, pawarSurveyCulturalAwareness2025}. When interacting with users from underrepresented regions such as South-East Asia (SEA), the LLMs may generate responses that are insensitive, irrelevant, or otherwise premised on Western cultural norms. Additionally, the Western bias presents the risk of flattening global cultural diversity. Therefore, improving the cultural awareness and understanding of LLMs has gained increasing research attention.

As the fourth populous country in the world, Indonesia has been historically under-represented in NLP research  \citep{ajiOneCountry7002022}. Indonesia is also one of the most ethnically and culturally diverse countries, with 600 to 1200 ethnic groups in the country, depending on the classification method \citep{bpsProfileEthnicGroups2024}. 
In recent years, there has been an intensifying effort to improve the availability of NLP resources for Indonesia. This includes development of benchmarks, such as \citet{kotoLargeLanguageModels2023} and \citet{ kotoIndoCultureExploringGeographically2024}, as well as a consolidation and standardization of disparate Indonesian datasets, as part of the SEACrowd initiative to facilitate usage of the datasets in research \citep{loveniaSEACrowdMultilingualMultimodal2024}.

One attractive source of cultural knowledge that has been largely overlooked is social science\footnote{For brevity, in this paper "social science" refers to both humanities and social sciences.} publications produced locally, which  likely contain studies into local cultures from a native perspective. We present a novel text dataset of journal article passages, \corpusname, which is created from Indonesian social science journals indexed in the Directory of Open Access Journals\footnote{\url{https://doaj.org/}}, and demonstrate its effectiveness on the Indonesian cultural benchmark, IndoCulture \citep{kotoIndoCultureExploringGeographically2024}.

On top of the dataset, the present study devises an effective technique to inject the cultural knowledge into LLMs. Inspired by previous research that retrieval may be more suitable for injecting specialized knowledge into LLMs than finetuning \citep{ovadiaFineTuningRetrievalComparing2024, soudaniFineTuningVs2024}, we propose to employ \corpusname{} in retrieval-augmented generation (RAG). First, we extract the factual statements regarding Indonesian culture in the journal articles. This is to prevent other types of text in the articles from interfering with the RAG process. During retrieval, following \citet{gaoPreciseZeroShotDense2023}, for each question the LLM is prompted to generate a hypothetical answer, which is then used as an informative key for retrieval.

We evaluate on the IndoCulture benchmark \citep{kotoIndoCultureExploringGeographically2024}, which covers diverse cultures in eleven Indonesian provinces.

The proposed recipe results in strong performance gains over baselines. The best model achieves an accuracy of 79.3\%, 2.9 percentage points higher than the previous SOTA. We carefully verify the effectiveness of all components of the proposed recipe using ablation experiments.

The contributions of this paper are as follows:
\begin{enumerate}[itemsep=0pt]
    \item We present a novel text dataset created from carefully parsed Indonesian social science publications, \corpusname, which contains Indonesian cultural knowledge. 
    \item We demonstrate a technique for injecting  Indonesian cultural knowledge from the journal articles into the LLMs. Facts extracted from the papers serves as values to be retrieved whereas hypothetical documents generated from the target question serve as the query.  
    \item We set a new SOTA of 81.4\% on IndoCulture by using RAG with a mixed corpus of Indonesian Wikipedia and extracted facts from journals, highlighting the potential of our corpus in complementing more common sources of knowledge.
\end{enumerate}

\section{Related Work}

\textbf{NLP Corpora of Academic Papers.} Academic publications in science and engineering fields have been included in NLP corpora. For example, papers from ArXiv and PubMed have been included in open-source datasets such as The Pile \citep{gaoPile800GBDataset2020}.  Beyond the STEM fields, S2ORC \citep{loS2ORCSemanticScholar2020} covered more academic disciplines by collecting papers from Semantic Scholar.  A small minority of papers contained in S2ORC are from social studies subjects such as Sociology, History, and Art; however, the dataset is limited to English-language papers. 

OpenMSD \citep{gaoOpenMSDMultilingualScientific2024} is a multilingual dataset of scientific papers, including papers from Southeast Asia (SEA). Nevertheless, the dataset is primarily intended for scientific document similarity measurement, and the data distribution is still heavily skewed towards STEM subjects. To our knowledge, there have been no previous corpora that focus on social science journal articles designed for the task of cultural understanding in Southeast Asia.

\vspace{6pt}
\noindent \textbf{Retrieval-Augmented Generation.} Pioneered by \citet{lewisRetrievalaugmentedGenerationKnowledgeintensive2020}, retrieval-augmented generation (RAG) is a technique for augmenting the internal knowledge of an LLM by retrieving documents from an external database and placing the retrieved documents in the LLM decision context.
Current implementation of RAG generally involves three steps: indexing, retrieval, and generation \citep{gaoRetrievalAugmentedGenerationLarge2024}. In the indexing step, text chunks are encoded into vector representations using an embedding model, and the vector representations are collected into a vector database. In the retrieval step, the same embedding model is used to encode a user query, and the similarity scores between the query vector and the vectors in the database are calculated. A predefined number \textit{D} of documents with the highest similarity scores is subsequently added to the prompt that will be given to the LLM, as context to the user query. In the generation step, the LLM is instructed to answer the new prompt containing both the context and the original query.

Much research has been conducted on RAG since the technique's introduction.  There has been research into improving the retrieval stage \citep{gaoPreciseZeroShotDense2023,zhuKnowledgeGraphGuidedRetrieval2025, laitenbergerStrongerBaselinesRetrievalAugmented2025}, instruction finetuning of the LLM to make more effective use of the retrieved documents \citep{liuRAGInstructBoostingLLMs2025, bhushanSystematicKnowledgeInjection2025}, and developing new evaluation frameworks \citep{zhuRAGEvalScenarioSpecific2025}. Other works investigate interleaving RAG with multi-step reasoning to improve the LLM's reasoning capability \citep{liSearcho1AgenticSearchEnhanced2025, jiangRAGStarEnhancingDeliberative2025}, and applying RAG to improve LLM performance in various domains \citep{liEvaluatingPerformanceRAG2025, wuMedicalGraphRAG2025}.

\vspace{6pt}
\noindent \textbf{Application of RAG for LLM Cultural Understanding.} There have been few studies into the use of RAG in cultural context. In the review of  \citet{pawarSurveyCulturalAwareness2025} regarding prior works on LLM cultural awareness,  it is reported that training-free methods to improve LLM cultural awareness have historically focused on prompting techniques. For example, \citet{chengMarkedPersonasUsing2023} prompted LLMs to generate personas of various demographic groups to investigate stereotypes contained in the LLMs; \citet{alkhamissiInvestigatingCulturalAlignment2024} showed that in Egyptian cultural context,  grounding LLM-generated personas using a framework derived from ethnographic studies improved the alignment of LLM responses and human participants' responses.

More recently, \citet{utamiFacilitatingAboriginalPerinatal2025} utilized RAG with a corpus of Indigenous Australian health information to create a culturally-sensitive chatbot in the context of mental health of Aboriginal mothers in Australia. Closer to the present work, 
\citet{leeEvaluatingCulturalKnowledge2025} developed a benchmark for Hakka culture, intended to test an LLM's capability across six aspects: remembering, understanding, applying, analyzing, evaluating, and creating. They showed that RAG with a corpus constructed primarily from Hakka-language Wikipedia leads to higher performance on their benchmark over a no-retrieval baseline. Nevertheless,  their study is focused on creating their benchmark instead of developing a dataset for improving the LLM performance, and they focused a single cultural group. Our work is intended to cover diverse cultural groups within Indonesia. 

\vspace{6pt}
\noindent \textbf{Computational Understanding of Indonesian Culture.} 
Early works on this topic focuses on LLMs' understanding of Indonesian language \citep{wilieIndoNLUBenchmarkResources2020, mahendraIndoNLINaturalLanguage2021}. 
More recently, research has focused more on LLM's reasoning ability related to Indonesian culture.  New benchmarks include testing the LLMs on Indonesian exam questions from primary school to university entrance levels  \citep{kotoLargeLanguageModels2023}, on Indonesian terminology, language nuances, and culture of Jakarta \citep{wibowoCOPALIDIndonesianLanguage2024}, as well as on 
human- and LLM- generated questions about general Indonesian culture \citep{putriCanLLMGenerate2024}.  The current most comprehensive benchmark, IndoCulture, evaluates LLMs'  understanding of Indonesian culture across eleven provinces, hence capturing the regional cultural diversity \citep{kotoIndoCultureExploringGeographically2024}.

\begin{figure}[t]
\centering
    \includegraphics[width=.48\textwidth]{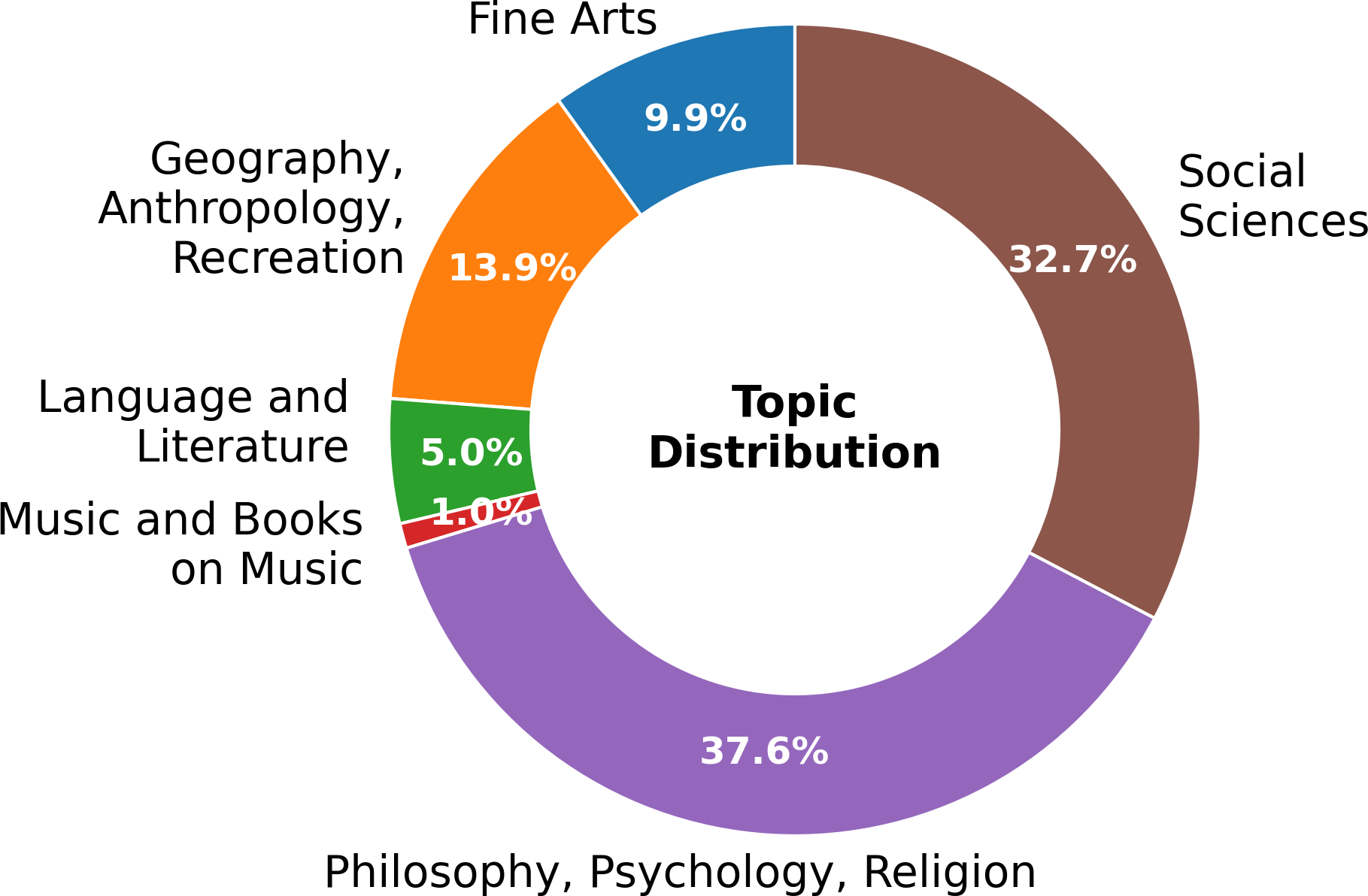}
    \caption{The proportions of social science topics covered by the \corpusname{} dataset. }
\label{fig:topic-proportions}
\end{figure}

\begin{figure}[t]
\centering
    \includegraphics[width=.48\textwidth]{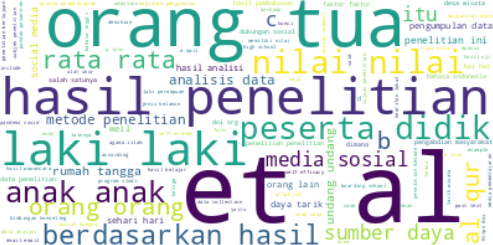}
    \caption{A word cloud of the frequent phrases in the academic text extracted from the regions labeled with main title, section title, abstract, text, and list. }
\label{fig:full-text-word-cloud}
\end{figure}

\begin{figure}[t]
\centering
    \includegraphics[width=.48\textwidth]{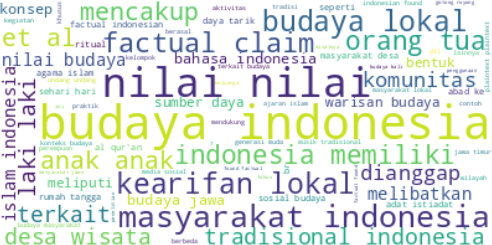}
    \caption{A word cloud of the frequent phrases in the text of the cultural facts extracted.}
\label{fig:fact-word-cloud}
\end{figure}

\section{Methodology}

\subsection{Creating the \corpusname{} Dataset}

We crawled Indonesian journals with Creative Commons licenses\footnote{CC-BY, CC-BY-SA, CC-BY-NC, and CC-BY-NC-SA} indexed in the Directory of Open Access Journals. From January to February 2025, we downloaded the pdf articles from all available online issues, yielding a total of  21,500 pdf files. The topic of journals, covering various social science topics, such as anthropology, ethics, and linguistic theory. We show a donut chart of the main category proportions in Figure \ref{fig:topic-proportions} and leave the complete ontology of topics to Appendix \ref{sec:ontology_doaj}.  

To facilitate downstream application of the dataset, we need to convert the collected pdf articles into plain text. However, one challenge we face is the complex layouts of academic publications, containing headers, footnotes, diagrams, tables, and single- or double-column text; simply extracting text line by line will mix text from different regions and disrupt the semantic meaning. To prevent this, we computationally identify the page layout, dividing each page into text regions and classifying them by function. 

Empirically, we find that off-the-shelf page layout detection systems to be insufficient for our purposes, as their label set is not designed for social science publications and they are not trained on Indonesian text. As a result, we redesigned the label space, annotated our own training data, and finetuned the network. The new label space contains four region labels from PubLayNet \citep{zhongPubLayNetLargestDataset2019}: text, list, table, and figure, as well as four additional labels that we created: main title, section title, abstract, and caption. 

We finetune an object detection network of LayoutLMv3 \citep{huangLayoutLMv3PretrainingDocument2022}. The network employs the LayoutLM backbone \cite{xu2020layoutlm}, which is a BERT-like text-vision bimodal Transformer, and a feature pyramid network \cite{lin2017feature-pyramid} for feature extraction, and Cascade R-CNN \citep{caiCascadeRCNNDelving2018} for region detection. The network is trained on medical publications in English from PubMed Central, creating distribution shift from Indonesian social science publications. Thus, we manually annotated the layout of 500 pages from \corpusname, including bounding boxes and classes of each box, and finetuned the Cascade R-CNN classification and regression heads accordingly. After finetuning, we attain a mean average precision (mAP) of 91.8\%.

For our purposes, we keep text from the following detected regional bounding boxes: main title, abstract, text, section title, or list were selected. We extract the text from the bounding boxes using the library PyMuPDF\footnote{\url{https://github.com/pymupdf/PyMuPDF}} and assemble them in the same order as they appear in the pdf file. To  exclude the bibliography, we remove text after the section title ``bibliography'', ``references'', or their equivalent in Bahasa Indonesia. This yields about 212 million tokens from 21,374 articles from 151 journals.

\subsection{Cultural Facts Extraction from Academic Text}

For computational understanding of Indonesian cultural practices, we are primarily interested in facts widely recognized among social scientists. However, social science publications often contain idiosyncratic opinions of authors or novel insights that have not yet reached consensus. During retrieval and generation, the existence of those text may mislead the RAG system. Therefore, we use an LLM (Sailor2-20B-Chat) to extract the facts related to Indonesian culture from the journal text. The LLM prompt used and an example output can be found in Appendices \ref{sec:prompt_fact_extraction} and \ref{sec:example_fact_extraction_result}, respectively.

Before fact extraction, we split the academic text into approximately 650,000 chunks of roughly three paragraphs each. All facts extracted from one chunk are merged together, forming one textual entry.  
The resulting dataset contains approximately 102,000 entries of Indonesian cultural facts and a total of 15 million tokens. The token yield ratio is 7.1\%. The relatively low ratio stems from the fact that much academic text does not describe cultural facts. For example, some journal articles may discuss statistical procedure at length. 

We visualize the text before and after the fact extraction step using word clouds in Fig. \ref{fig:full-text-word-cloud} and \ref{fig:fact-word-cloud}. Before fact extraction, the most frequent phrases are research-related expressions such as "hasil penelitian" (results of the study) and generic expressions like "laki laki'' (male). After fact extraction, phrases like ``kearifan lokal'' (local wisdom) and ``nilai nilai'' (values) become more frequent, suggesting we indeed capture local cultural values and practices.

\subsection{RAG with Hypothetical Documents}
We now describe the RAG pipeline. As the result of the fact extraction step, we have access to a number of textual entries regarding Indonesian culture. Given a question about Indonesian culture and a few answer choices, we retrieve at the level of textual entries. We recognize the existence of a distributional gap between the question, and the facts that can be used to answer the question. Thus, instead of using the question as the query, we follow the example of \citet{gaoPreciseZeroShotDense2023} to use as the query a hypothetical document that is more similar to the factual statements to be retrieved. 

More specifically, we prompt the LLM being tested to generate a synthetic document that might provide the answer. Note that the synthetic document is not used to answer the question, only to retrieve relevant facts. Therefore, we do not expect the synthetic document to be factually correct, only that it is distributionally similar to the correct factual entry that we want to retrieve. After that, we apply an embedding model to convert the synthetic document to a query vector. We then retrieve the textual entries with the highest cosine similarity and place them in the LLM context, from which the LLM answers the target question.

\section{Experiments}

In this section we present two sets of experiments:
\begin{enumerate}[itemsep=0pt]
    \item RAG with our corpus of extracted facts from social science journals
    \item RAG with a mixed corpus of Indonesian Wikipedia + extracted facts from social science journals
\end{enumerate}

For each set of experiments we present the main results of RAG performance on the IndoCulture benchmark, the ablation studies, and additional experiments.

\subsection{Setup} 

\textbf{Indonesian Culture Benchmark.} Our proposed method was tested on the IndoCulture benchmark \citep{kotoIndoCultureExploringGeographically2024}. This commonsense reasoning benchmark contains 2,429 questions designed to test an LLM's understanding of various cultural topics, ranging from food to religious holidays, across eleven Indonesian provinces. IndoCulture is currently the most comprehensive benchmark on Indonesian culture. The multiple-choice question (MCQ) format with province context was chosen our experiments. The prompt format used is provided in Appendix~\ref{sec:prompt_benchmark}.

\vspace{6pt}
\noindent \textbf{MCQ Evaluation.} Following \citet{kotoLargeLanguageModels2023} and \citet{kotoIndoCultureExploringGeographically2024}, for each question in IndoCulture we obtain the probabilities for the first generated token and select the probabilities that correspond to the answer choices (A, B, C). The answer choice with the highest probability is taken as the model's answer for that question.

\vspace{6pt}
\noindent \textbf{LLMs Employed.} We applied our proposed method on recent models that are specifically developed for Southeast Asian languages, including SeaLLMs-v3 \citep{zhangSeaLLMs3Open2025}, Sailor2 \citep{douSailor2SailingSouthEast2025}, and SEA-LION v4 \citep{ngSEALIONSoutheastAsian2025}.  We experimented with both the base pretrained models and finetuned chat models where available.

The state-of-the-art (SOTA) performance reported previously for IndoCulture is 76.4, using Sailor2-20B model \citep{douSailor2SailingSouthEast2025}. The regular versions of Sailor2 models have relatively short context lengths of 4096; to accommodate the large number of tokens from all the retrieved passages, for Sailor2 models, we conducted RAG experiments with the long-context variants. 

\vspace{6pt}
\noindent \textbf{RAG Details.} The raw journal articles were chunked using the recursive text splitter from LangChain\footnote{\url{https://docs.langchain.com/oss/python/integrations/splitters/recursive_text_splitter}}  with a chunk size of 1600. To encode texts into vector representations, BGE-M3 \citep{chenM3EmbeddingMultiLingualityMultiFunctionality2024} was chosen as the embedding model due to its multilingual capabilities. The resulting vector embeddings from the text chunks are indexed using FAISS on GPU \citep{johnsonBillionScaleSimilaritySearch2019}.

In generating both the facts from journal article chunks and the hypothetical documents that may answer the benchmark question, we used a temperature of 0.5, top-p sampling with p = 0.9, and no top-k sampling. The vLLM library \citep{kwonEfficientMemoryManagement2023} was used in both generation tasks for efficiency.

\subsection{RAG Extracted Facts from Social Science Journals}

\begin{table}
\centering
\begin{tabular}{@{}lcc@{}}
\toprule
Base LLM & \textbf{No RAG} & \textbf{D=20} \\
 \midrule
SEALLMs-v3-7B & 54.6 & 61.3 \\
SEALLMs- v3-7B-Chat & 60.6 & 65.3 \\
Sailor2-L-8B & 64.2 & 74.5 \\
Sailor2-L-8B-Chat & 70.5 & 73.9 \\
Sailor2-L-20B & 72.1 & 75.7 \\
Sailor2-L-20B-Chat & 75.4 & \textbf{79.3} \\
Qwen-SEA-LION-v4-32B-IT & 70.9 & 75.5 \\
\bottomrule
\end{tabular}
\caption{Zero-shot accuracy on IndoCulture using RAG with extracted facts and hypothetical document queries. The results in the No RAG column were obtained by directly prompting the model with the benchmark questions, without any additional context. We use 20 retrieved passages in the LLM decision context.}
\label{tab:full_ext_facts_journal}
\end{table}

Table \ref{tab:full_ext_facts_journal} presents the performance of RAG with the corpus of extracted facts from social science journals. On all the LLMs tested, retrieval from \corpusname{} results in considerable performance improvement over the no-retrieval baseline. In particular, the best score of 79.3 achieved by Sailor2-L-20B-Chat with RAG is better than the previously reported SOTA of 76.4 \citep{douSailor2SailingSouthEast2025}. 

The performance gain starts to be observed when the number of retrieved passages \textit{D } is one. The improvement in performance generally increases with increasing \textit{D}, although as noted by \citet{ovadiaFineTuningRetrievalComparing2024} the optimal number of retrieved passages may be both model- and task-dependent. To avoid tuning the hyperparameter on the test set, we simply report the performance at D=20.

\begin{table}[t]
\centering
\small
\begin{tabular}{@{}lccc@{}}
\toprule
\multicolumn{4}{c}{\textbf{Base   case : RAG with journal extracted facts}}               \\
\multicolumn{4}{c}{\textbf{Ablation   case :  RAG with raw journal text chunks}}           \\
\midrule
\multicolumn{1}{c}{\textbf{}} & \textbf{Base} & \textbf{Ablation} & \textbf{B-A} \\
\midrule
SEALLMs-v3-7B               & 60.3    & 60.4        & -0.1    \\
SEALLMs- v3-7B-Chat         & 63.9    & 62.4        & +1.5    \\
Sailor2-L-8B                & 73.4    & 70.7        & +2.7    \\
Sailor2-L-8B-Chat           & 73.5    & 74.5        & -1.0    \\
Sailor2-L-20B               & 75.1    & 73.1        & +2.0    \\
Sailor2-L-20B-Chat          & 78.6    & 78.0        & +0.6    \\
Qwen-SEA-LION-v4-32B-IT     & 74.7    & 73.5        & +1.2    \\ \bottomrule
\end{tabular}
\caption{Average change in RAG performance on IndoCulture when using the corpus of journal extracted facts, over an ablation case of using the corpus of raw journal texts.} 
\label{tab:ext_facts_journal_vs_raw}
\end{table}

\vspace{6pt}
\noindent \textbf{Ablation: Fact Extraction from Scientific Texts}
In this ablation study, we analyze the effectiveness of the fact extraction step. Table \ref{tab:ext_facts_journal_vs_raw}  demonstrates that for most models, RAG using the corpus of journal extracted facts yields better performance than RAG with the corpus of raw journal texts. 

The observed performance gain suggests that presenting the cultural knowledge in as a collection of facts is indeed important. The training data for the tested models include Wikipedia articles \citep{zhangSeaLLMs3Open2025, douSailor2SailingSouthEast2025, ngSEALIONSoutheastAsian2025}; the format of a Wikipedia article can be seen as a series of facts. As such, converting the academic style of the social science journal articles into a format the LLMs are already familiar with could help the LLMs in utilizing the knowledge content.

\begin{table}[t]
\centering
\small
\begin{tabular}{@{}lccc@{}}
\toprule
\multicolumn{4}{c}{\textbf{Base   case : RAG with hypothetical documents}}           \\
\multicolumn{4}{c}{\textbf{Ablation   case :  RAG with no hypothetical documents}}         \\
\midrule
\multicolumn{1}{c}{\textbf{}} & \textbf{Base} & \textbf{Ablation} & \textbf{B-A} \\
\midrule
SEALLMs-v3-7B                 & 60.3          & 59.6              & +0.7         \\
SEALLMs- v3-7B-Chat           & 63.9          & 64.0              & -0.1         \\
Sailor2-L-8B                  & 73.4          & 72.1              & +1.3         \\
Sailor2-L-8B-Chat             & 73.5          & 71.5              & +2.0         \\
Sailor2-L-20B                 & 75.1          & 73.2              & +1.9         \\
Sailor2-L-20B-Chat            & 78.6          & 77.2              & +1.4         \\
Qwen-SEA-LION-v4-32B-IT       & 74.7          & 73.4              & +1.3         \\ \bottomrule
\end{tabular}
\caption{Average change in RAG performance on IndoCulture when using model-generated hypothetical documents as the retrieval queries, over an ablation case of using the IndoCulture benchmark questions as the queries. }
\label{tab:hyde_vs_no_hyde}
\end{table}

\vspace{6pt}
\noindent \textbf{Ablation: Hypothetical Documents as Query.}
We conducted an ablation study to investigate the impact of using the model-generated hypothetical answers as the retrieval queries. 
The results (Table \ref{tab:hyde_vs_no_hyde}) show that in most models tested, using hypothetical documents as the retrieval queries outperforms using the IndoCulture questions as the queries. This is in line with the result of  \citet{gaoPreciseZeroShotDense2023}. The observed trend in Table \ref{tab:hyde_vs_no_hyde} that hypothetical document generation is more applicable for the stronger models is also in line with their observation.

\vspace{6pt}
\noindent \textbf{Ablation: Alternative Textual Units for Retrieval.}
The raw journal texts includes text chunks that contain no cultural knowledge, such as discussion of statistical procedures. That is why we specifically extract cultural facts from the journal text before using them in RAG. However, it is possible that the cultural fact extraction step is overly aggressive and remove necessary context for the facts, which may mislead the RAG system. 

In this ablation study, we try to retain the immediate context of the extracted cultural facts by keeping the entire raw text chunks around the facts. If a textual chunk does not contain any fact, it is discarded. We call the resulting corpus the filtered raw corpus.
We present a comparison of RAG performance with the extracted facts corpus and this filtered raw corpus in Table \ref{tab:ext_facts_journal_vs_filtered_raw}. The observed result indicates that for most models, the additional context confuses more than it clarifies. 

\begin{table}[t]
\centering
\small
\begin{tabular}{@{}lccc@{}}
\toprule
\multicolumn{4}{c}{\textbf{Base   case : RAG with journal extracted facts}}                \\
\multicolumn{4}{c}{\textbf{Ablation   case : with raw texts of the   extracted facts}} \\
\midrule
\multicolumn{1}{c}{\textbf{}}    & \textbf{Base}    & \textbf{Ablation}   & \textbf{B-A}   \\
\midrule
SEALLMs-v3-7B                    & 60.3             & 60.4                & -0.1           \\
SEALLMs- v3-7B-Chat              & 63.9             & 63.4                & +0.5           \\
Sailor2-L-8B                     & 73.4             & 71.0                & +2.4           \\
Sailor2-L-8B-Chat                & 73.5             & 74.2                & -0.7           \\
Sailor2-L-20B                    & 75.1             & 73.3                & +1.8           \\
Sailor2-L-20B-Chat               & 78.6             & 78.5                & +0.1           \\
Qwen-SEA-LION-v4-32B-IT          & 74.7             & 73.9                & +0.8           \\ \bottomrule
\end{tabular}
\caption{Average change in RAG performance on IndoCulture when using the extracted facts from the journal text chunks as the retrieval corpus, over an ablation case of using the corresponding raw text chunks as the corpus. 
}
\label{tab:ext_facts_journal_vs_filtered_raw}
\end{table}

Continuing this line of inquiry, we investigate the relative importance of the extracted facts during the retrieval step and the generation step. In the retrieval step, it may be easier to discriminate the relevant passages from the less relevant ones when the embeddings are made from the extracted facts rather than the raw text chunks. Alternatively, in the generation step, the format of the passages added as context to the LLM prompt may be important. We conducted an experiment in which the embeddings of the extracted facts were used for similarity calculations with the hypothetical answers, but the passages added to the LLM context were the corresponding raw texts of the extracted facts. 

\begin{table}[ht]
\centering
\small
\begin{tabular}{@{}lccc@{}}
\toprule 
\multicolumn{4}{c}{\textbf{Base   case : Ext. facts for retrieval and generation}} \\
\multicolumn{4}{c}{\textbf{Ablation   case : Ext. facts for retrieval only}}       \\
\midrule
\multicolumn{1}{c}{\textbf{}}  & \textbf{Base}  & \textbf{Ablation} & \textbf{B-A} \\
\midrule
SEALLMs-v3-7B                  & 60.3           & 60.5              & -0.2         \\
SEALLMs- v3-7B-Chat            & 63.9           & 63.1              & +0.8         \\
Sailor2-L-8B                   & 73.4           & 71.5              & +1.9         \\
Sailor2-L-8B-Chat              & 73.5           & 74.6              & -1.1         \\
Sailor2-L-20B                  & 75.1           & 73.1              & +2.0         \\
Sailor2-L-20B-Chat             & 78.6           & 78.5              & +0.1         \\
Qwen-SEA-LION-v4-32B-IT        & 74.7           & 73.9              & +0.8         \\ \bottomrule
\end{tabular}
\caption{Average change in RAG performance on IndoCulture when the extracted facts are used for both the embedding similarity calculation and as passages added to LLM context, over an ablation case of using the raw text chunks of the extracted facts as the context passages. 
}
\label{tab:crossrag_journal}
\end{table}

From Table \ref{tab:crossrag_journal}, for most models using the extracted facts for both embedding similarity calculation and as context passages still outperforms using the extracted facts only for embedding similarity calculation. This indicates that using the extracted facts throughout the RAG pipeline is advantageous.

\subsection{RAG with Mixture of Extracted Journal Facts and Wikipedia}

To further explore the potential of our extracted facts corpus for RAG application, we propose to append our corpus to Indonesian Wikipedia text. We hypothesize that scholarly publications from social science journal may contain different kinds of knowledge that complement Wikipedia. For example, the knowledge contained in Wikipedia may be more widely known, while the journals may incorporate more knowledge about cultural minorities or ancient practices. 

The Wikipedia corpus was created from a dump of Indonesian-language Wikipedia dated 20 August 2025. To improve the relevance of the articles for downstream retrieval application, the articles included in the corpus are restricted to those containing "Indonesia" or the name of an Indonesian province in the main text. The articles are chunked with the same settings as those used for the journal articles. The resulting corpus contains 184,000 passages and 103 million tokens. The chunked Wikipedia articles were combined with the extracted facts from social science journals, and the mixed corpus was indexed as a single vector database.

\begin{table}
\centering
\begin{tabular}{@{}lcc@{}}   
\toprule
\textbf{Base LLM} & \textbf{No RAG} & \textbf{D=20} \\
\midrule
SEALLMs-v3-7B & 54.6 & 64.1 \\
SEALLMs- v3-7B-Chat & 60.6 & 66.3 \\
Sailor2-L-8B & 64.2 & 74.3 \\
Sailor2-L-8B-Chat & 70.5 & 77.2 \\
Sailor2-L-20B & 72.1 & 78.0 \\
Sailor2-L-20B-Chat & 75.4 & \textbf{81.4} \\
Qwen-SEA-LION-v4-32B-IT & 70.9 & 79.5 \\
\bottomrule
\end{tabular}
\caption{Zero-shot accuracy on IndoCulture using RAG on both cultural facts extracted from \corpusname{} and Indonesian Wikipedia. The results in the column labeled "No RAG" were obtained by directly prompting the model with the benchmark questions, without any additional context.}
\label{tab:full_results_mixed}
\end{table}

\vspace{6pt}
\noindent \textbf{Main Result.} The results of RAG with the mixed corpus shown in Table \ref{tab:full_results_mixed} show even stronger performance gains over the no-retrieval baseline. With the help of retrieval from the mixed corpus, the best results from four models outperform the previous SOTA of 76.4 on IndoCulture. The score of 81.4 obtained using Sailor2-L-20B-Chat sets a new SOTA for the benchmark.

\vspace{6pt}
\noindent \textbf{Ablation: Effects of Journal Text.}
The goal of this ablation study is to evaluate the impact of adding our corpus of extracted facts to a corpus of Wikipedia texts.

\begin{table}[t]
\centering
\small
\begin{tabular}{@{}lccc@{}}
\toprule 
\multicolumn{4}{c}{\textbf{Base   case : RAG with Wikipedia + journal ext. facts}} \\
\multicolumn{4}{c}{\textbf{Ablation   case : RAG with Wikipedia only}}             \\
\midrule
\multicolumn{1}{c}{\textbf{}}  & \textbf{Base}  & \textbf{Ablation} & \textbf{B-A} \\
\midrule
SEALLMs-v3-7B                  & 63.6           & 62.1              & +1.5         \\
SEALLMs- v3-7B-Chat            & 66.3           & 65.4              & +0.9         \\
Sailor2-L-8B                   & 73.8           & 72.9              & +0.9         \\
Sailor2-L-8B-Chat              & 76.5           & 76.0              & +0.5         \\
Sailor2-L-20B                  & 77.2           & 76.9              & +0.3         \\
Sailor2-L-20B-Chat             & 80.4           & 79.4              & +1.0         \\
Qwen-SEA-LION-v4-32B-IT        & 78.2           & 76.7              & +1.5         \\ \bottomrule
\end{tabular}
\caption{Average change in RAG performance on IndoCulture when using the mixed corpus of journal extracted facts and Wikipedia texts, over an ablation case of using only the Wikipedia texts. 
}
\label{tab:mixed_vs_wikipedia}
\end{table}

Table \ref{tab:mixed_vs_wikipedia} shows that RAG using the mixed corpus outperforms RAG using the corpus of only Wikipedia texts for all models tested. This observation suggests that our specialized corpus of extracted facts from social science journals can well complement a corpus created from common sources such as Wikipedia. Correspondingly, the observed results also support our hypothesis that social science journals may contain cultural knowledge that is distinct from that already captured in Wikipedia. Exactly how the knowledge content of the two sources differ is an interesting avenue for investigation in future research.

\begin{table}[t]
\centering
\small
\begin{tabular}{@{}lccc@{}}
\toprule 
\multicolumn{4}{c}{\textbf{Case   1 : RAG with Wikipedia extracted facts}}             \\
\multicolumn{4}{c}{\textbf{Case   2: RAG with raw Wikipedia}}                          \\
\midrule
\multicolumn{1}{c}{\textbf{}} & \textbf{Case 1} & \textbf{Case   2} & \textbf{1   - 2} \\
\midrule
SEALLMs-v3-7B                 & 62.6            & 62.1              & +0.5             \\
SEALLMs- v3-7B-Chat           & 66.1            & 65.4              & +0.7             \\
Sailor2-L-8B                  & 73.4            & 72.9              & +0.5             \\
Sailor2-L-8B-Chat             & 73.8            & 76.0              & -2.2             \\
Sailor2-L-20B                 & 76.1            & 76.9              & -0.8             \\
Sailor2-L-20B-Chat            & 78.3            & 79.4              & -1.1             \\
Qwen-SEA-LION-v4-32B-IT       & 76.2            & 76.7              & -0.5             \\ \bottomrule
\end{tabular}

\caption{Average change in RAG performance on IndoCulture when using extracted facts from Wikipedia chunks as the retrieval corpus, over a case of using the raw Wikipedia text chunks. 
}
\label{tab:ext_facts_wiki_vs_raw_wiki}
\end{table}

\vspace{6pt}
\noindent \textbf{Ablation: Fact Extraction on Wikipedia Text.}
To test whether the fact extraction can help regardless of original text format, we apply the fact extraction prompt on Wikipedia text chunks. We conduct RAG experiments using a corpus of extracted facts from Wikipedia and using a corpus of raw Wikipedia texts.

Table \ref{tab:ext_facts_wiki_vs_raw_wiki} shows that for the stronger models, RAG using a corpus of extracted facts from Wikipedia leads to worse results than using the raw Wikipedia corpus. However, the weaker models such as SEALLMs-v3-7B and Sailor2-L-8B benefit from the additional fact extraction step.

A possible reason is that, as the result of its editing process, Wikipedia is already quite clean and contains mostly widely recognized facts. The stronger models are already capable of utilizing cultural knowledge from raw Wikipedia. 
Further trimming it down could lose contextual information or introduce errors. This result is similar to the finding of   \citet{laitenbergerStrongerBaselinesRetrievalAugmented2025}, who reported that retrieving original passages leads to better RAG performance than retrieving generated summaries.

This experiment therefore highlights the pertinence of applying the fact extraction step to our corpus of Indonesian social science journal articles. As an illustration, fact extraction results from two passages about traditional Indonesian snacks are shown in Appendix \ref{sec:example_fact_extraction_result} (from journal article) and Appendix \ref{sec:example_ext_fact_wikipedia} (from Wikipedia article). In Appendix \ref{sec:example_fact_extraction_result}, the argumentative style of the original journal passage regarding the history of the dish is converted into shorter factual statements regarding the origin and ingredients of the dish. In contrast, as shown in Appendix \ref{sec:example_ext_fact_wikipedia}, the extracted factual statement from the Wikipedia passage is remarkably similar to the original passage. Some information regarding the ingredients of the dish has also not been included in the extracted factual statement.

\section{Conclusion}

In this paper we explore the utilization of Indonesian social science journals to inject cultural knowledge into LLMs in the understanding of Indonesian culture. We present a novel text dataset of journal article passages, created from 151 open-source Indonesian social science journals. We use a strong LLM to extract facts related to Indonesian culture from the raw journal text passages. We subsequently use the resulting corpus of extracted facts for retrieval-augmented generation. 
We show that our proposed method results in strong performance gains over the no-retrieval baseline on the IndoCulture benchmark. Additionally, by combining our corpus with Indonesian Wikipedia, our best RAG performance on IndoCulture sets a new SOTA accuracy of 81.4\%.

\section*{Limitations}

The journal articles that we collected are written exclusively in Indonesian or English. Meanwhile, Indonesia has more than 700 spoken languages \citep{ajiOneCountry7002022}. As such, our journal corpus may not fully capture the richness of Indonesian cultural traditions.

Furthermore, this paper focuses on improving an LLM's knowledge of Indonesian cultural practices. We have not evaluated whether our method can allow an LLM to understand "deeper" aspects of culture, such as nuanced understanding of Indonesian language or culturally appropriate responses in conversational contexts.

\section*{Acknowledgments}
This research is supported by the RIE2025 Industry Alignment Fund – Industry Collaboration Projects (IAF-ICP) (Award I2301E0026), administered by A*STAR, as well as supported by Alibaba Group and NTU Singapore through Alibaba-NTU Global e-Sustainability CorpLab (ANGEL).

\bibliography{reftex}

@inproceedings{adilazuardaMeasuringModelingCulture2024,
  title = {Towards {{Measuring}} and {{Modeling}} ``{{Culture}}'' in {{LLMs}}: {{A Survey}}},
  shorttitle = {Towards {{Measuring}} and {{Modeling}} ``{{Culture}}'' in {{LLMs}}},
  booktitle = {Proceedings of the 2024 {{Conference}} on {{Empirical Methods}} in {{Natural Language Processing}}},
  author = {Adilazuarda, Muhammad Farid and Mukherjee, Sagnik and Lavania, Pradhyumna and Singh, Siddhant Shivdutt and Aji, Alham Fikri and O'Neill, Jacki and Modi, Ashutosh and Choudhury, Monojit},
  year = 2024,
  pages = {15763--15784},
  publisher = {Association for Computational Linguistics},
  address = {Miami, Florida, USA},
  doi = {10.18653/v1/2024.emnlp-main.882},
  urldate = {2025-12-12},
  langid = {english}
}

@inproceedings{ajiOneCountry7002022,
  title = {One {{Country}}, 700+ {{Languages}}: {{NLP Challenges}} for {{Underrepresented Languages}} and {{Dialects}} in {{Indonesia}}},
  shorttitle = {One {{Country}}, 700+ {{Languages}}},
  booktitle = {Proceedings of the 60th {{Annual Meeting}} of the {{Association}} for {{Computational Linguistics}} ({{Volume}} 1: {{Long Papers}})},
  author = {Aji, Alham Fikri and Winata, Genta Indra and Koto, Fajri and Cahyawijaya, Samuel and Romadhony, Ade and Mahendra, Rahmad and Kurniawan, Kemal and Moeljadi, David and Prasojo, Radityo Eko and Baldwin, Timothy and Lau, Jey Han and Ruder, Sebastian},
  editor = {Muresan, Smaranda and Nakov, Preslav and Villavicencio, Aline},
  year = 2022,
  month = may,
  pages = {7226--7249},
  publisher = {Association for Computational Linguistics},
  address = {Dublin, Ireland},
  doi = {10.18653/v1/2022.acl-long.500},
  urldate = {2025-12-23}
}

@inproceedings{alkhamissiInvestigatingCulturalAlignment2024,
  title = {Investigating {{Cultural Alignment}} of {{Large Language Models}}},
  booktitle = {Proceedings of the 62nd {{Annual Meeting}} of the {{Association}} for {{Computational Linguistics}} ({{Volume}} 1: {{Long Papers}})},
  author = {AlKhamissi, Badr and ElNokrashy, Muhammad and Alkhamissi, Mai and Diab, Mona},
  editor = {Ku, Lun-Wei and Martins, Andre and Srikumar, Vivek},
  year = 2024,
  month = aug,
  pages = {12404--12422},
  publisher = {Association for Computational Linguistics},
  address = {Bangkok, Thailand},
  doi = {10.18653/v1/2024.acl-long.671},
  urldate = {2025-12-27}
}

@inproceedings{bhushanSystematicKnowledgeInjection2025,
  title = {Systematic {{Knowledge Injection}} into {{Large Language Models}} via {{Diverse Augmentation}} for {{Domain-Specific RAG}}},
  booktitle = {Findings of the {{Association}} for {{Computational Linguistics}}: {{NAACL}} 2025},
  author = {Bhushan, Kushagra and Nandwani, Yatin and Khandelwal, Dinesh and Gupta, Sonam and Pandey, Gaurav and Raghu, Dinesh and Joshi, Sachindra},
  editor = {Chiruzzo, Luis and Ritter, Alan and Wang, Lu},
  year = 2025,
  month = apr,
  pages = {5922--5943},
  publisher = {Association for Computational Linguistics},
  address = {Albuquerque, New Mexico},
  doi = {10.18653/v1/2025.findings-naacl.329},
  urldate = {2026-01-06},
  isbn = {979-8-89176-195-7}
}

@misc{bpsProfileEthnicGroups2024,
  title = {Profile of {{Ethnic Groups}} and {{Regional Language Diversity Results}} of the 2020 {{Population Census Long Form}}},
  author = {{BPS}},
  year = 2024,
  urldate = {2026-01-04},
  howpublished = {\url{https://www.bps.go.id/en/publication/2024/12/12/6feb932e24186429686fb57b/profile-of-ethnic-groups-and-regional-language-diversity-results-of-the-2020-population-census-long-form.html}},
  langid = {english}
}

@inproceedings{caiCascadeRCNNDelving2018,
  title = {Cascade {{R-CNN}}: {{Delving Into High Quality Object Detection}}},
  shorttitle = {Cascade {{R-CNN}}},
  booktitle = {2018 {{IEEE}}/{{CVF Conference}} on {{Computer Vision}} and {{Pattern Recognition}}},
  author = {Cai, Zhaowei and Vasconcelos, Nuno},
  year = 2018,
  month = jun,
  pages = {6154--6162},
  issn = {2575-7075},
  doi = {10.1109/CVPR.2018.00644},
  urldate = {2026-01-04}
}

@inproceedings{caoAssessingCrossCulturalAlignment2023,
  title = {Assessing {{Cross-Cultural Alignment}} between {{ChatGPT}} and {{Human Societies}}: {{An Empirical Study}}},
  shorttitle = {Assessing {{Cross-Cultural Alignment}} between {{ChatGPT}} and {{Human Societies}}},
  booktitle = {Proceedings of the {{First Workshop}} on {{Cross-Cultural Considerations}} in {{NLP}} ({{C3NLP}})},
  author = {Cao, Yong and Zhou, Li and Lee, Seolhwa and Cabello, Laura and Chen, Min and Hershcovich, Daniel},
  editor = {Dev, Sunipa and Prabhakaran, Vinodkumar and Adelani, David Ifeoluwa and Hovy, Dirk and Benotti, Luciana},
  year = 2023,
  month = may,
  pages = {53--67},
  publisher = {Association for Computational Linguistics},
  address = {Dubrovnik, Croatia},
  doi = {10.18653/v1/2023.c3nlp-1.7},
  urldate = {2025-12-27}
}

@inproceedings{chengMarkedPersonasUsing2023,
  title = {Marked {{Personas}}: {{Using Natural Language Prompts}} to {{Measure Stereotypes}} in {{Language Models}}},
  shorttitle = {Marked {{Personas}}},
  booktitle = {Proceedings of the 61st {{Annual Meeting}} of the {{Association}} for {{Computational Linguistics}} ({{Volume}} 1: {{Long Papers}})},
  author = {Cheng, Myra and Durmus, Esin and Jurafsky, Dan},
  editor = {Rogers, Anna and {Boyd-Graber}, Jordan and Okazaki, Naoaki},
  year = 2023,
  month = jul,
  pages = {1504--1532},
  publisher = {Association for Computational Linguistics},
  address = {Toronto, Canada},
  doi = {10.18653/v1/2023.acl-long.84},
  urldate = {2025-12-27}
}

@inproceedings{chenM3EmbeddingMultiLingualityMultiFunctionality2024,
  title = {M3-{{Embedding}}: {{Multi-Linguality}}, {{Multi-Functionality}}, {{Multi-Granularity Text Embeddings Through Self-Knowledge Distillation}}},
  shorttitle = {M3-{{Embedding}}},
  booktitle = {Findings of the {{Association}} for {{Computational Linguistics}}: {{ACL}} 2024},
  author = {Chen, Jianlyu and Xiao, Shitao and Zhang, Peitian and Luo, Kun and Lian, Defu and Liu, Zheng},
  editor = {Ku, Lun-Wei and Martins, Andre and Srikumar, Vivek},
  year = 2024,
  month = aug,
  pages = {2318--2335},
  publisher = {Association for Computational Linguistics},
  address = {Bangkok, Thailand},
  doi = {10.18653/v1/2024.findings-acl.137},
  urldate = {2025-12-12}
}

@misc{douSailor2SailingSouthEast2025,
  title = {Sailor2: {{Sailing}} in {{South-East Asia}} with {{Inclusive Multilingual LLMs}}},
  shorttitle = {Sailor2},
  author = {Dou, Longxu and Liu, Qian and Zhou, Fan and Chen, Changyu and Wang, Zili and Jin, Ziqi and Liu, Zichen and Zhu, Tongyao and Du, Cunxiao and Yang, Penghui and Wang, Haonan and Liu, Jiaheng and Zhao, Yongchi and Feng, Xiachong and Mao, Xin and Yeung, Man Tsung and Pipatanakul, Kunat and Koto, Fajri and Thu, Min Si and Kydl{\'i}{\v c}ek, Hynek and Liu, Zeyi and Lin, Qunshu and Sripaisarnmongkol, Sittipong and {Sae-Khow}, Kridtaphad and Thongchim, Nirattisai and Konkaew, Taechawat and Borijindargoon, Narong and Dao, Anh and Maneegard, Matichon and Artkaew, Phakphum and Yong, Zheng-Xin and Nguyen, Quan and Phatthiyaphaibun, Wannaphong and Tran, Hoang H. and Zhang, Mike and Chen, Shiqi and Pang, Tianyu and Du, Chao and Wan, Xinyi and Lu, Wei and Lin, Min},
  year = 2025,
  month = feb,
  number = {arXiv:2502.12982},
  eprint = {2502.12982},
  primaryclass = {cs},
  publisher = {arXiv},
  doi = {10.48550/arXiv.2502.12982},
  urldate = {2025-12-12},
  archiveprefix = {arXiv},
  langid = {english}
}

@article{elstyPENELUSURANSEJARAHFILOSOFI2020,
  title = {{{PENELUSURAN SEJARAH}}, {{FILOSOFI DAN BUDAYA MAKAN KUE GEPLAK KHAS BETAWI}}},
  author = {Elsty, Kezia and Nahdlah, Zayyini},
  year = 2020,
  month = dec,
  journal = {Jurnal Pariwisata Pesona},
  volume = {5},
  number = {2},
  pages = {69--75},
  issn = {2541-5859},
  doi = {10.26905/jpp.v5i1.4745},
  urldate = {2026-01-04},
  langid = {english}
}

@inproceedings{gaoOpenMSDMultilingualScientific2024,
  title = {{{OpenMSD}}: {{Towards Multilingual Scientific Documents Similarity Measurement}}},
  shorttitle = {{{OpenMSD}}},
  booktitle = {Proceedings of the 2024 {{Joint International Conference}} on {{Computational Linguistics}}, {{Language Resources}} and {{Evaluation}} ({{LREC-COLING}} 2024)},
  author = {Gao, Yang and Ma, Ji and Korotkov, Ivan and Hall, Keith and Alon, Dana and Metzler, Donald},
  editor = {Calzolari, Nicoletta and Kan, Min-Yen and Hoste, Veronique and Lenci, Alessandro and Sakti, Sakriani and Xue, Nianwen},
  year = 2024,
  month = may,
  pages = {12467--12480},
  publisher = {{ELRA and ICCL}},
  address = {Torino, Italia},
  urldate = {2025-12-28}
}

@misc{gaoPile800GBDataset2020,
  title = {The {{Pile}}: {{An 800GB Dataset}} of {{Diverse Text}} for {{Language Modeling}}},
  shorttitle = {The {{Pile}}},
  author = {Gao, Leo and Biderman, Stella and Black, Sid and Golding, Laurence and Hoppe, Travis and Foster, Charles and Phang, Jason and He, Horace and Thite, Anish and Nabeshima, Noa and Presser, Shawn and Leahy, Connor},
  year = 2020,
  month = dec,
  number = {arXiv:2101.00027},
  eprint = {2101.00027},
  primaryclass = {cs},
  publisher = {arXiv},
  doi = {10.48550/arXiv.2101.00027},
  urldate = {2025-12-28},
  archiveprefix = {arXiv}
}

@inproceedings{gaoPreciseZeroShotDense2023,
  title = {Precise {{Zero-Shot Dense Retrieval}} without {{Relevance Labels}}},
  booktitle = {Proceedings of the 61st {{Annual Meeting}} of the {{Association}} for {{Computational Linguistics}} ({{Volume}} 1: {{Long Papers}})},
  author = {Gao, Luyu and Ma, Xueguang and Lin, Jimmy and Callan, Jamie},
  year = 2023,
  pages = {1762--1777},
  publisher = {Association for Computational Linguistics},
  address = {Toronto, Canada},
  doi = {10.18653/v1/2023.acl-long.99},
  urldate = {2025-12-12},
  langid = {english}
}

@misc{gaoRetrievalAugmentedGenerationLarge2024,
  title = {Retrieval-{{Augmented Generation}} for {{Large Language Models}}: {{A Survey}}},
  shorttitle = {Retrieval-{{Augmented Generation}} for {{Large Language Models}}},
  author = {Gao, Yunfan and Xiong, Yun and Gao, Xinyu and Jia, Kangxiang and Pan, Jinliu and Bi, Yuxi and Dai, Yi and Sun, Jiawei and Wang, Meng and Wang, Haofen},
  year = 2024,
  month = mar,
  number = {arXiv:2312.10997},
  eprint = {2312.10997},
  primaryclass = {cs},
  publisher = {arXiv},
  doi = {10.48550/arXiv.2312.10997},
  urldate = {2025-12-25},
  archiveprefix = {arXiv}
}

@inproceedings{huangLayoutLMv3PretrainingDocument2022,
  title = {{{LayoutLMv3}}: {{Pre-training}} for {{Document AI}} with {{Unified Text}} and {{Image Masking}}},
  shorttitle = {{{LayoutLMv3}}},
  booktitle = {Proceedings of the 30th {{ACM International Conference}} on {{Multimedia}}},
  author = {Huang, Yupan and Lv, Tengchao and Cui, Lei and Lu, Yutong and Wei, Furu},
  year = 2022,
  month = oct,
  series = {{{MM}} '22},
  pages = {4083--4091},
  publisher = {Association for Computing Machinery},
  address = {New York, NY, USA},
  doi = {10.1145/3503161.3548112},
  urldate = {2025-12-11},
  isbn = {978-1-4503-9203-7}
}

@inproceedings{jiangRAGStarEnhancingDeliberative2025,
  title = {{{RAG-Star}}: {{Enhancing Deliberative Reasoning}} with {{Retrieval Augmented Verification}} and {{Refinement}}},
  shorttitle = {{{RAG-Star}}},
  booktitle = {Proceedings of the 2025 {{Conference}} of the {{Nations}} of the {{Americas Chapter}} of the {{Association}} for {{Computational Linguistics}}: {{Human Language Technologies}} ({{Volume}} 1: {{Long Papers}})},
  author = {Jiang, Jinhao and Chen, Jiayi and Li, Junyi and Ren, Ruiyang and Wang, Shijie and Zhao, Wayne Xin and Song, Yang and Zhang, Tao},
  editor = {Chiruzzo, Luis and Ritter, Alan and Wang, Lu},
  year = 2025,
  month = apr,
  pages = {7064--7074},
  publisher = {Association for Computational Linguistics},
  address = {Albuquerque, New Mexico},
  doi = {10.18653/v1/2025.naacl-long.361},
  urldate = {2026-01-06},
  isbn = {979-8-89176-189-6}
}

@article{johnsonBillionScaleSimilaritySearch2019,
  title = {Billion-{{Scale Similarity Search}} with {{GPUs}}},
  author = {Johnson, Jeff and Douze, Matthijs and J{\'e}gou, Herv{\'e}},
  year = 2019,
  journal = {IEEE Transactions on Big Data},
  volume = {7},
  number = {3},
  pages = {535--547},
  issn = {2332-7790},
  doi = {10.1109/TBDATA.2019.2921572},
  urldate = {2025-12-12}
}

@article{kotoIndoCultureExploringGeographically2024,
  title = {{{IndoCulture}}: {{Exploring Geographically Influenced Cultural Commonsense Reasoning Across Eleven Indonesian Provinces}}},
  shorttitle = {{{IndoCulture}}},
  author = {Koto, Fajri and Mahendra, Rahmad and Aisyah, Nurul and Baldwin, Timothy},
  year = 2024,
  journal = {Transactions of the Association for Computational Linguistics},
  volume = {12},
  pages = {1703--1719},
  publisher = {MIT Press},
  address = {Cambridge, MA},
  doi = {10.1162/tacl_a_00726},
  urldate = {2025-12-12}
}

@inproceedings{kotoLargeLanguageModels2023,
  title = {Large {{Language Models Only Pass Primary School Exams}} in {{Indonesia}}: {{A Comprehensive Test}} on {{IndoMMLU}}},
  shorttitle = {Large {{Language Models Only Pass Primary School Exams}} in {{Indonesia}}},
  booktitle = {Proceedings of the 2023 {{Conference}} on {{Empirical Methods}} in {{Natural Language Processing}}},
  author = {Koto, Fajri and Aisyah, Nurul and Li, Haonan and Baldwin, Timothy},
  year = 2023,
  pages = {12359--12374},
  publisher = {Association for Computational Linguistics},
  address = {Singapore},
  doi = {10.18653/v1/2023.emnlp-main.760},
  urldate = {2025-12-12},
  langid = {english}
}

@inproceedings{kwonEfficientMemoryManagement2023,
  title = {Efficient {{Memory Management}} for {{Large Language Model Serving}} with {{PagedAttention}}},
  booktitle = {Proceedings of the 29th {{Symposium}} on {{Operating Systems Principles}}},
  author = {Kwon, Woosuk and Li, Zhuohan and Zhuang, Siyuan and Sheng, Ying and Zheng, Lianmin and Yu, Cody Hao and Gonzalez, Joseph and Zhang, Hao and Stoica, Ion},
  year = 2023,
  month = oct,
  series = {{{SOSP}} '23},
  pages = {611--626},
  publisher = {Association for Computing Machinery},
  address = {New York, NY, USA},
  doi = {10.1145/3600006.3613165},
  urldate = {2025-12-11},
  isbn = {979-8-4007-0229-7}
}

@inproceedings{laitenbergerStrongerBaselinesRetrievalAugmented2025,
  title = {Stronger {{Baselines}} for {{Retrieval-Augmented Generation}} with {{Long-Context Language Models}}},
  booktitle = {Proceedings of the 2025 {{Conference}} on {{Empirical Methods}} in {{Natural Language Processing}}},
  author = {Laitenberger, Alex and Manning, Christopher D and Liu, Nelson F.},
  editor = {Christodoulopoulos, Christos and Chakraborty, Tanmoy and Rose, Carolyn and Peng, Violet},
  year = 2025,
  month = nov,
  pages = {32547--32557},
  publisher = {Association for Computational Linguistics},
  address = {Suzhou, China},
  doi = {10.18653/v1/2025.emnlp-main.1656},
  urldate = {2025-12-12},
  isbn = {979-8-89176-332-6}
}

@article{leeEvaluatingCulturalKnowledge2025,
  title = {Evaluating Cultural Knowledge Processing in Large Language Models: A Cognitive Benchmarking Framework Integrating Retrieval-Augmented Generation},
  shorttitle = {Evaluating Cultural Knowledge Processing in Large Language Models},
  author = {Lee, Hung-Shin and Chang, Chen-Chi and Chen, Ching-Yuan and Hsu, Yun-Hsiang},
  year = 2025,
  month = nov,
  journal = {The Electronic Library},
  pages = {1--22},
  issn = {0264-0473},
  doi = {10.1108/EL-04-2025-0136},
  urldate = {2025-12-28}
}

@inproceedings{lewisRetrievalaugmentedGenerationKnowledgeintensive2020,
  title = {Retrieval-Augmented Generation for Knowledge-Intensive {{NLP}} Tasks},
  booktitle = {Proceedings of the 34th {{International Conference}} on {{Neural Information Processing Systems}}},
  author = {Lewis, Patrick and Perez, Ethan and Piktus, Aleksandra and Petroni, Fabio and Karpukhin, Vladimir and Goyal, Naman and K{\"u}ttler, Heinrich and Lewis, Mike and Yih, Wen-tau and Rockt{\"a}schel, Tim and Riedel, Sebastian and Kiela, Douwe},
  year = 2020,
  month = dec,
  series = {{{NIPS}} '20},
  pages = {9459--9474},
  publisher = {Curran Associates Inc.},
  address = {Red Hook, NY, USA},
  urldate = {2025-12-11},
  isbn = {978-1-7138-2954-6}
}

@inproceedings{liEvaluatingPerformanceRAG2025,
  title = {Evaluating the {{Performance}} of {{RAG Methods}} for {{Conversational AI}} in the {{Airport Domain}}},
  booktitle = {Proceedings of the 2025 {{Conference}} of the {{Nations}} of the {{Americas Chapter}} of the {{Association}} for {{Computational Linguistics}}: {{Human Language Technologies}} ({{Volume}} 3: {{Industry Track}})},
  author = {Li, Yuyang and Kerbusch, Pjm and Pruim, Rhr and K{\"a}fer, Tobias},
  editor = {Chen, Weizhu and Yang, Yi and Kachuee, Mohammad and Fu, Xue-Yong},
  year = 2025,
  month = apr,
  pages = {794--808},
  publisher = {Association for Computational Linguistics},
  address = {Albuquerque, New Mexico},
  doi = {10.18653/v1/2025.naacl-industry.61},
  urldate = {2026-01-06},
  isbn = {979-8-89176-194-0}
}

@inproceedings{liSearcho1AgenticSearchEnhanced2025,
  title = {Search-O1: {{Agentic Search-Enhanced Large Reasoning Models}}},
  shorttitle = {Search-O1},
  booktitle = {Proceedings of the 2025 {{Conference}} on {{Empirical Methods}} in {{Natural Language Processing}}},
  author = {Li, Xiaoxi and Dong, Guanting and Jin, Jiajie and Zhang, Yuyao and Zhou, Yujia and Zhu, Yutao and Zhang, Peitian and Dou, Zhicheng},
  editor = {Christodoulopoulos, Christos and Chakraborty, Tanmoy and Rose, Carolyn and Peng, Violet},
  year = 2025,
  month = nov,
  pages = {5420--5438},
  publisher = {Association for Computational Linguistics},
  address = {Suzhou, China},
  doi = {10.18653/v1/2025.emnlp-main.276},
  urldate = {2026-01-06},
  isbn = {979-8-89176-332-6}
}

@inproceedings{liuRAGInstructBoostingLLMs2025,
  title = {{{RAG-Instruct}}: {{Boosting LLMs}} with {{Diverse Retrieval-Augmented Instructions}}},
  shorttitle = {{{RAG-Instruct}}},
  booktitle = {Proceedings of the 2025 {{Conference}} on {{Empirical Methods}} in {{Natural Language Processing}}},
  author = {Liu, Wanlong and Chen, Junying and Ji, Ke and Zhou, Li and Chen, Wenyu and Wang, Benyou},
  editor = {Christodoulopoulos, Christos and Chakraborty, Tanmoy and Rose, Carolyn and Peng, Violet},
  year = 2025,
  month = nov,
  pages = {3865--3888},
  publisher = {Association for Computational Linguistics},
  address = {Suzhou, China},
  doi = {10.18653/v1/2025.emnlp-main.192},
  urldate = {2026-01-06},
  isbn = {979-8-89176-332-6}
}

@inproceedings{loS2ORCSemanticScholar2020,
  title = {{{S2ORC}}: {{The Semantic Scholar Open Research Corpus}}},
  shorttitle = {{{S2ORC}}},
  booktitle = {Proceedings of the 58th {{Annual Meeting}} of the {{Association}} for {{Computational Linguistics}}},
  author = {Lo, Kyle and Wang, Lucy Lu and Neumann, Mark and Kinney, Rodney and Weld, Daniel},
  editor = {Jurafsky, Dan and Chai, Joyce and Schluter, Natalie and Tetreault, Joel},
  year = 2020,
  month = jul,
  pages = {4969--4983},
  publisher = {Association for Computational Linguistics},
  address = {Online},
  doi = {10.18653/v1/2020.acl-main.447},
  urldate = {2025-12-27}
}

@inproceedings{loveniaSEACrowdMultilingualMultimodal2024,
  title = {{{SEACrowd}}: {{A Multilingual Multimodal Data Hub}} and {{Benchmark Suite}} for {{Southeast Asian Languages}}},
  shorttitle = {{{SEACrowd}}},
  booktitle = {Proceedings of the 2024 {{Conference}} on {{Empirical Methods}} in {{Natural Language Processing}}},
  author = {Lovenia, Holy and Mahendra, Rahmad and Akbar, Salsabil Maulana and Miranda, Lester James V. and Santoso, Jennifer and Aco, Elyanah and Fadhilah, Akhdan and Mansurov, Jonibek and Imperial, Joseph Marvin and Kampman, Onno P. and Moniz, Joel Ruben Antony and Habibi, Muhammad Ravi Shulthan and Hudi, Frederikus and Montalan, Railey and Ignatius, Ryan and Lopo, Joanito Agili and Nixon, William and Karlsson, B{\"o}rje F. and Jaya, James and Diandaru, Ryandito and Gao, Yuze and Amadeus, Patrick and Wang, Bin and Cruz, Jan Christian Blaise and Whitehouse, Chenxi and Parmonangan, Ivan Halim and Khelli, Maria and Zhang, Wenyu and Susanto, Lucky and Ryanda, Reynard Adha and Hermawan, Sonny Lazuardi and Velasco, Dan John and Kautsar, Muhammad Dehan Al and Hendria, Willy Fitra and Moslem, Yasmin and Flynn, Noah and Adilazuarda, Muhammad Farid and Li, Haochen and Lee, Johanes and Damanhuri, R. and Sun, Shuo and Qorib, Muhammad Reza and Djanibekov, Amirbek and Leong, Wei Qi and Do, Quyet V. and Muennighoff, Niklas and Pansuwan, Tanrada and Putra, Ilham Firdausi and Xu, Yan and Chia, Tai Ngee and Purwarianti, Ayu and Ruder, Sebastian and Tjhi, William and Limkonchotiwat, Peerat and Aji, Alham Fikri and Keh, Sedrick and Winata, Genta Indra and Zhang, Ruochen and Koto, Fajri and Yong, Zheng-Xin and Cahyawijaya, Samuel},
  editor = {{Al-Onaizan}, Yaser and Bansal, Mohit and Chen, Yun-Nung},
  year = 2024,
  month = nov,
  pages = {5155--5203},
  publisher = {Association for Computational Linguistics},
  address = {Miami, Florida, USA},
  doi = {10.18653/v1/2024.emnlp-main.296},
  urldate = {2025-12-19}
}

@inproceedings{mahendraIndoNLINaturalLanguage2021,
  title = {{{IndoNLI}}: {{A Natural Language Inference Dataset}} for {{Indonesian}}},
  shorttitle = {{{IndoNLI}}},
  booktitle = {Proceedings of the 2021 {{Conference}} on {{Empirical Methods}} in {{Natural Language Processing}}},
  author = {Mahendra, Rahmad and Aji, Alham Fikri and Louvan, Samuel and Rahman, Fahrurrozi and Vania, Clara},
  editor = {Moens, Marie-Francine and Huang, Xuanjing and Specia, Lucia and Yih, Scott Wen-tau},
  year = 2021,
  month = nov,
  pages = {10511--10527},
  publisher = {Association for Computational Linguistics},
  address = {Online and Punta Cana, Dominican Republic},
  doi = {10.18653/v1/2021.emnlp-main.821},
  urldate = {2026-01-06}
}

@misc{ngSEALIONSoutheastAsian2025,
  title = {{{SEA-LION}}: {{Southeast Asian Languages}} in {{One Network}}},
  shorttitle = {{{SEA-LION}}},
  author = {Ng, Raymond and Nguyen, Thanh Ngan and Huang, Yuli and Tai, Ngee Chia and Leong, Wai Yi and Leong, Wei Qi and Yong, Xianbin and Ngui, Jian Gang and Susanto, Yosephine and Cheng, Nicholas and Rengarajan, Hamsawardhini and Limkonchotiwat, Peerat and Hulagadri, Adithya Venkatadri and Teng, Kok Wai and Tong, Yeo Yeow and Siow, Bryan and Teo, Wei Yi and Lau, Wayne and Tan, Choon Meng and Ong, Brandon and Ong, Zhi Hao and Montalan, Jann Railey and Chan, Adwin and Antonyrex, Sajeban and Lee, Ren and Choa, Esther and {Tat-Wee}, David Ong and Liu, Bing Jie Darius and Tjhi, William Chandra and Cambria, Erik and Teo, Leslie},
  year = 2025,
  month = oct,
  number = {arXiv:2504.05747},
  eprint = {2504.05747},
  primaryclass = {cs},
  publisher = {arXiv},
  doi = {10.48550/arXiv.2504.05747},
  urldate = {2025-12-12},
  archiveprefix = {arXiv},
  langid = {english}
}

@inproceedings{ovadiaFineTuningRetrievalComparing2024,
  title = {Fine-{{Tuning}} or {{Retrieval}}? {{Comparing Knowledge Injection}} in {{LLMs}}},
  shorttitle = {Fine-{{Tuning}} or {{Retrieval}}?},
  booktitle = {Proceedings of the 2024 {{Conference}} on {{Empirical Methods}} in {{Natural Language Processing}}},
  author = {Ovadia, Oded and Brief, Menachem and Mishaeli, Moshik and Elisha, Oren},
  year = 2024,
  pages = {237--250},
  publisher = {Association for Computational Linguistics},
  address = {Miami, Florida, USA},
  doi = {10.18653/v1/2024.emnlp-main.15},
  urldate = {2025-12-12},
  langid = {english}
}

@article{pawarSurveyCulturalAwareness2025,
  title = {Survey of {{Cultural Awareness}} in {{Language Models}}: {{Text}} and {{Beyond}}},
  shorttitle = {Survey of {{Cultural Awareness}} in {{Language Models}}},
  author = {Pawar, Siddhesh and Park, Junyeong and Jin, Jiho and Arora, Arnav and Myung, Junho and Yadav, Srishti and Haznitrama, Faiz Ghifari and Song, Inhwa and Oh, Alice and Augenstein, Isabelle},
  year = 2025,
  month = sep,
  journal = {Computational Linguistics},
  volume = {51},
  number = {3},
  pages = {907--1004},
  issn = {0891-2017, 1530-9312},
  doi = {10.1162/COLI.a.14},
  urldate = {2025-12-12},
  langid = {english}
}

@inproceedings{putriCanLLMGenerate2024,
  title = {Can {{LLM Generate Culturally Relevant Commonsense QA Data}}? {{Case Study}} in {{Indonesian}} and {{Sundanese}}},
  shorttitle = {Can {{LLM Generate Culturally Relevant Commonsense QA Data}}?},
  booktitle = {Proceedings of the 2024 {{Conference}} on {{Empirical Methods}} in {{Natural Language Processing}}},
  author = {Putri, Rifki Afina and Haznitrama, Faiz Ghifari and Adhista, Dea and Oh, Alice},
  editor = {{Al-Onaizan}, Yaser and Bansal, Mohit and Chen, Yun-Nung},
  year = 2024,
  month = nov,
  pages = {20571--20590},
  publisher = {Association for Computational Linguistics},
  address = {Miami, Florida, USA},
  doi = {10.18653/v1/2024.emnlp-main.1145},
  urldate = {2026-01-06}
}

@inproceedings{soudaniFineTuningVs2024,
  title = {Fine {{Tuning}} vs. {{Retrieval Augmented Generation}} for {{Less Popular Knowledge}}},
  booktitle = {Proceedings of the 2024 {{Annual International ACM SIGIR Conference}} on {{Research}} and {{Development}} in {{Information Retrieval}} in the {{Asia Pacific Region}}},
  author = {Soudani, Heydar and Kanoulas, Evangelos and Hasibi, Faegheh},
  year = 2024,
  month = dec,
  pages = {12--22},
  publisher = {ACM},
  address = {Tokyo Japan},
  doi = {10.1145/3673791.3698415},
  urldate = {2025-12-12},
  isbn = {979-8-4007-0724-7},
  langid = {english}
}

@inproceedings{utamiFacilitatingAboriginalPerinatal2025,
  title = {Facilitating {{Aboriginal Perinatal Mental Health Information Access}} with a {{Retrieval-Augmented LLM-based Chatbot}}},
  booktitle = {2025 47th {{Annual International Conference}} of the {{IEEE Engineering}} in {{Medicine}} and {{Biology Society}} ({{EMBC}})},
  author = {Utami, Made Srinitha Millinia and Kwok, Wai Hang and Kotz, Jayne and Walker, Roz and Wang, Guanjin and Marriott, Rhonda},
  year = 2025,
  month = jul,
  pages = {1--7},
  issn = {2694-0604},
  doi = {10.1109/EMBC58623.2025.11254329},
  urldate = {2025-12-28}
}

@inproceedings{wibowoCOPALIDIndonesianLanguage2024,
  title = {{{COPAL-ID}}: {{Indonesian Language Reasoning}} with {{Local Culture}} and {{Nuances}}},
  shorttitle = {{{COPAL-ID}}},
  booktitle = {Proceedings of the 2024 {{Conference}} of the {{North American Chapter}} of the {{Association}} for {{Computational Linguistics}}: {{Human Language Technologies}} ({{Volume}} 1: {{Long Papers}})},
  author = {Wibowo, Haryo and Fuadi, Erland and Nityasya, Made and Prasojo, Radityo Eko and Aji, Alham},
  editor = {Duh, Kevin and Gomez, Helena and Bethard, Steven},
  year = 2024,
  month = jun,
  pages = {1404--1422},
  publisher = {Association for Computational Linguistics},
  address = {Mexico City, Mexico},
  doi = {10.18653/v1/2024.naacl-long.77},
  urldate = {2026-01-06}
}

@inproceedings{wilieIndoNLUBenchmarkResources2020,
  title = {{{IndoNLU}}: {{Benchmark}} and {{Resources}} for {{Evaluating Indonesian Natural Language Understanding}}},
  shorttitle = {{{IndoNLU}}},
  booktitle = {Proceedings of the 1st {{Conference}} of the {{Asia-Pacific Chapter}} of the {{Association}} for {{Computational Linguistics}} and the 10th {{International Joint Conference}} on {{Natural Language Processing}}},
  author = {Wilie, Bryan and Vincentio, Karissa and Winata, Genta Indra and Cahyawijaya, Samuel and Li, Xiaohong and Lim, Zhi Yuan and Soleman, Sidik and Mahendra, Rahmad and Fung, Pascale and Bahar, Syafri and Purwarianti, Ayu},
  editor = {Wong, Kam-Fai and Knight, Kevin and Wu, Hua},
  year = 2020,
  month = dec,
  pages = {843--857},
  publisher = {Association for Computational Linguistics},
  address = {Suzhou, China},
  doi = {10.18653/v1/2020.aacl-main.85},
  urldate = {2026-01-06}
}

@inproceedings{wuMedicalGraphRAG2025,
  title = {Medical {{Graph RAG}}: {{Evidence-based Medical Large Language Model}} via {{Graph Retrieval-Augmented Generation}}},
  shorttitle = {Medical {{Graph RAG}}},
  booktitle = {Proceedings of the 63rd {{Annual Meeting}} of the {{Association}} for {{Computational Linguistics}} ({{Volume}} 1: {{Long Papers}})},
  author = {Wu, Junde and Zhu, Jiayuan and Qi, Yunli and Chen, Jingkun and Xu, Min and Menolascina, Filippo and Jin, Yueming and Grau, Vicente},
  editor = {Che, Wanxiang and Nabende, Joyce and Shutova, Ekaterina and Pilehvar, Mohammad Taher},
  year = 2025,
  month = jul,
  pages = {28443--28467},
  publisher = {Association for Computational Linguistics},
  address = {Vienna, Austria},
  doi = {10.18653/v1/2025.acl-long.1381},
  urldate = {2026-01-06},
  isbn = {979-8-89176-251-0}
}

@inproceedings{zhangSeaLLMs3Open2025,
  title = {{{SeaLLMs}} 3: {{Open Foundation}} and {{Chat Multilingual Large Language Models}} for {{Southeast Asian Languages}}},
  shorttitle = {{{SeaLLMs}} 3},
  booktitle = {Proceedings of the 2025 {{Conference}} of the {{Nations}} of the {{Americas Chapter}} of the {{Association}} for {{Computational Linguistics}}: {{Human Language Technologies}} ({{System Demonstrations}})},
  author = {Zhang, Wenxuan and Chan, Hou Pong and Zhao, Yiran and Aljunied, Mahani and Wang, Jianyu and Liu, Chaoqun and Deng, Yue and Hu, Zhiqiang and Xu, Weiwen and Chia, Yew Ken and Li, Xin and Bing, Lidong},
  editor = {Dziri, Nouha and Ren, Sean (Xiang) and Diao, Shizhe},
  year = 2025,
  month = apr,
  pages = {96--105},
  publisher = {Association for Computational Linguistics},
  address = {Albuquerque, New Mexico},
  doi = {10.18653/v1/2025.naacl-demo.10},
  urldate = {2025-12-12},
  isbn = {979-8-89176-191-9}
}

@misc{zhongPubLayNetLargestDataset2019,
  title = {{{PubLayNet}}: Largest Dataset Ever for Document Layout Analysis},
  shorttitle = {{{PubLayNet}}},
  author = {Zhong, Xu and Tang, Jianbin and Yepes, Antonio Jimeno},
  year = 2019,
  month = aug,
  number = {arXiv:1908.07836},
  eprint = {1908.07836},
  primaryclass = {cs},
  publisher = {arXiv},
  doi = {10.48550/arXiv.1908.07836},
  urldate = {2026-01-04},
  archiveprefix = {arXiv}
}

@inproceedings{zhuKnowledgeGraphGuidedRetrieval2025,
  title = {Knowledge {{Graph-Guided Retrieval Augmented Generation}}},
  booktitle = {Proceedings of the 2025 {{Conference}} of the {{Nations}} of the {{Americas Chapter}} of the {{Association}} for {{Computational Linguistics}}: {{Human Language Technologies}} ({{Volume}} 1: {{Long Papers}})},
  author = {Zhu, Xiangrong and Xie, Yuexiang and Liu, Yi and Li, Yaliang and Hu, Wei},
  editor = {Chiruzzo, Luis and Ritter, Alan and Wang, Lu},
  year = 2025,
  month = apr,
  pages = {8912--8924},
  publisher = {Association for Computational Linguistics},
  address = {Albuquerque, New Mexico},
  doi = {10.18653/v1/2025.naacl-long.449},
  urldate = {2026-01-06},
  isbn = {979-8-89176-189-6}
}

@inproceedings{zhuRAGEvalScenarioSpecific2025,
  title = {{{RAGEval}}: {{Scenario Specific RAG Evaluation Dataset Generation Framework}}},
  shorttitle = {{{RAGEval}}},
  booktitle = {Proceedings of the 63rd {{Annual Meeting}} of the {{Association}} for {{Computational Linguistics}} ({{Volume}} 1: {{Long Papers}})},
  author = {Zhu, Kunlun and Luo, Yifan and Xu, Dingling and Yan, Yukun and Liu, Zhenghao and Yu, Shi and Wang, Ruobing and Wang, Shuo and Li, Yishan and Zhang, Nan and Han, Xu and Liu, Zhiyuan and Sun, Maosong},
  editor = {Che, Wanxiang and Nabende, Joyce and Shutova, Ekaterina and Pilehvar, Mohammad Taher},
  year = 2025,
  month = jul,
  pages = {8520--8544},
  publisher = {Association for Computational Linguistics},
  address = {Vienna, Austria},
  doi = {10.18653/v1/2025.acl-long.418},
  urldate = {2026-01-06},
  isbn = {979-8-89176-251-0}
}

@inproceedings{xu2020layoutlm,
  title={Layoutlm: Pre-training of text and layout for document image understanding},
  author={Xu, Yiheng and Li, Minghao and Cui, Lei and Huang, Shaohan and Wei, Furu and Zhou, Ming},
  booktitle={Proceedings of the 26th ACM SIGKDD international conference on knowledge discovery \& data mining},
  pages={1192--1200},
  year={2020}
}

@inproceedings{lin2017feature-pyramid,
  title={Feature pyramid networks for object detection},
  author={Lin, Tsung-Yi and Doll{\'a}r, Piotr and Girshick, Ross and He, Kaiming and Hariharan, Bharath and Belongie, Serge},
  booktitle={Proceedings of the IEEE conference on computer vision and pattern recognition},
  pages={2117--2125},
  year={2017}
}

\appendix

\newpage
\section{Risks}
\label{sec:risks}
A cultural tradition associated with a cultural group may not be practiced by all members of that cultural group. Users of our corpus should keep this in mind to avoid stereotyping the members of a cultural group. Additionally,  research findings published in the social science journals regarding particular cultural practices or social phenomena may be time-dependent. As such, future users of our dataset should take care to verify that such information are still applicable.

\section{Prompt for Fact Extraction}
\label{sec:prompt_fact_extraction}

The following prompt is used to extract facts related to Indonesian culture from a chunk of journal article text. The text passage is placed in the [DOCUMENT] field.

\begin{tcolorbox}[title=Prompt for Fact Extraction]
\small
Extract all factual claims related to Indonesian culture from the following passage. Enclose your response within <factual\_claims> and </factual\_claims> tags. 
Write the factual claims in Indonesian. 
If you cannot find any factual claims related to Indonesian culture, write 'No relevant factual claims found’.
\vspace{1mm}
\vspace{1mm}

PASSAGE:
\vspace{1mm}

[DOCUMENT]
\vspace{1mm}

OUTPUT: <factual\_claims></factual\_claims>
\end{tcolorbox}

\newpage
\section{Example Fact Extraction Result from Journal}
\label{sec:example_fact_extraction_result}

The following box provides an example of the resulting factual statements extracted from a raw text passage in our journal dataset. The text passage is taken from a paper by \citet{elstyPENELUSURANSEJARAHFILOSOFI2020} about Kue Geplak Betawi, which is a traditional dish.

\begin{tcolorbox}[title=Example Fact Extraction Result from Journal]
\small
Original text passage:
\vspace{1mm}

Sejarah Kue Geplak Betawi
\vspace{1mm}

Bila dilihat dari berbagai pendekatan, setidaknya ada lima perspektif untuk mengenal asal Kue  Geplak khas Betawi. Pendekatan pertama dapat dilihat dari asal kue ini tercipta. Saputra (2019)  menjelaskan saat ini tidak ada dokumen tertulis dan tidak diketahui persisnya kapan kue ini tercipta.  Namun keberadaan kue ini dapat dikaitkan dengan keberadaan ekosistem dengan segala hasil bumi di  dalamnya. Makanan dan lingkungan merupakan hal yang tidak dipisahkan karena pembuatan makanan  sebagian besar dari ketersedaan bahan pangan yang disediakan oleh lingkungan yang dalam hal ini  disebut ekosistem pada suatu tempat di mana aktivitas keseharian dan kehidupan suatu masyarakat  berlangsung (Untari, 2016). Ketersediaan bahan pada ekosistem yang ada mempengaruhi pemikiran  nenek moyang untuk mengolah hasi bumi yang tercatat juga pada relief-relief Candi Borobudur di mana  kegiatan terkait kuliner sudah ada sejak abad ke-8 yang ditandai dengan adanya relief kegiatan memasak  di dapur atau masyarakat yang mengangkut padi. Kreatifitas masyarakat juga yang akhirnya melahirkan  sebuah makanan khususnya kue. Asal Kue Geplak Betawi dapat ditelusuri dari membedah bahan-bahan  dari kue ini salah satunya beras.  Bahan lainnya adalah kelapa parut sangrai, larutan gula, daun pandan,  dan daun jeruk. Kue Geplak Betawi ini dapat ditemui pada masyarakat Betawi pinggiran. Penggunaan  bahan hasil bumi padi yang diolah menjadi kue Geplak Betawi juga menjadi upaya untuk menjadikan  bahan tersebut untuk memiliki manfaat lebih yakni menjadi cemilan kue.

\hrulefill
\vspace{1mm}

Extracted factual statements:
\vspace{1mm}

Kue Geplak adalah makanan khas Betawi.
\vspace{1mm}

Asal-usulnya terkait dengan ekosistem dan hasil bumi, terutama beras dan kelapa.
\vspace{1mm}

Kue Geplak Betawi menggunakan bahan seperti beras, kelapa parut sangrai, larutan gula, daun pandan, dan daun jeruk.
\vspace{1mm}

Kue ini ditemukan di masyarakat Betawi pinggiran, mencerminkan kreativitas lokal dalam pengolahan hasil bumi.
\vspace{1mm}

\end{tcolorbox}

English translations of the original passage and the extracted facts are provided in the following box.

\begin{tcolorbox}[title=English translation from ChatGPT, after skip=10pt]
\small
Original text passage:
\vspace{1mm}

History of Betawi Geplak Cake
\vspace{1mm}

When viewed from various approaches, there are at least five perspectives for understanding the origins of Betawi-style Geplak cake. The first approach looks at how this cake came into existence. Saputra (2019) explains that, to date, there are no written documents, and it is not known exactly when this cake was first created. However, its existence can be linked to the presence of an ecosystem and all the natural resources within it. Food and the environment are inseparable, because food production largely depends on the availability of food ingredients provided by the environment, which in this context is referred to as the ecosystem of a place where daily activities and the life of a community take place (Untari, 2016). The availability of materials within an ecosystem influenced the thinking of ancestors in processing agricultural products, a fact also recorded in the reliefs of Borobudur Temple, where culinary-related activities have existed since the 8th century. This is indicated by reliefs depicting cooking activities in kitchens or people transporting rice. The creativity of the community ultimately gave rise to various foods, especially cakes. The origin of Betawi Geplak cake can be traced by examining its ingredients, one of which is rice. Other ingredients include toasted grated coconut, sugar syrup, pandan leaves, and kaffir lime leaves. Betawi Geplak cake can be found among Betawi communities living on the outskirts. The use of rice-based agricultural products processed into Betawi Geplak cake also represents an effort to give these ingredients added value by turning them into snack foods.

\hrulefill
\vspace{1mm}

Extracted factual statements:
\vspace{1mm}

Geplak cake is a traditional Betawi food.
\vspace{1mm}

Its origin is connected to the ecosystem and natural resources, especially rice and coconut.\vspace{1mm}

Betawi Geplak cake uses ingredients such as rice, toasted grated coconut, sugar syrup, pandan leaves, and kaffir lime leaves.\vspace{1mm}

This cake is found among Betawi communities on the outskirts, reflecting local creativity in processing natural resources.
\vspace{1mm}

\end{tcolorbox}

\newpage
\section{Example Fact Extraction Result from Wikipedia}
\label{sec:example_ext_fact_wikipedia}

The following box provides an example of the resulting factual statements extracted from an Indonesian Wikipedia passage. The article is titled "Geplak", included a Wikipedia dump dated 20 August 2025. Geplak is distinct from Kue Geplak Betawi in Appendix \ref{sec:example_fact_extraction_result}, although they share some characteristics.

\begin{tcolorbox}[title=Example Fact Extraction Result from Wikipedia]
\small
Original text passage:
\vspace{1mm}

Geplak adalah penganan yang dibuat dari adonan kelapa parut (ampas kelapa) dicampur gula dan vanili, ada yang dicampuri durian, sirsak, atau nangka.
\vspace{1mm}

Geplak merupakan penganan tradisional khas Jawa yang berasal dari kabupaten Bantul, Daerah Istimewa Yogyakarta. Terdapat pula geplak yang dibuat dari waluh. Industri geplak umumnya dapat ditemui di daerah Kabupaten Bantul, Daerah Istimewa Yogyakarta, yang kebanyakan diusahakan oleh industri rumah tangga. Selanjutnya jenis penganan ini berkembang meluas akibat permintaan pasar dan diusahakan tidak hanya di sekitar Daerah Istimewa Yogyakarta akan tetapi juga di seluruh Nusantara.

\hrulefill
\vspace{1mm}

Extracted factual statements:
\vspace{1mm}

Geplak adalah penganan tradisional khas Jawa dari Kabupaten Bantul, Daerah Istimewa Yogyakarta.
Industri geplak umumnya diusahakan oleh industri rumah tangga di Bantul dan telah berkembang ke seluruh Nusantara.

\end{tcolorbox}

\begin{tcolorbox}[title=English Translation from ChatGPT]
\small
Original text passage:
\vspace{1mm}

Geplak is a snack made from grated coconut (coconut pulp) mixed with sugar and vanilla, and sometimes flavored with durian, soursop, or jackfruit.
\vspace{1mm}

Geplak is a traditional snack originally from Bantul Regency, Special Region of Yogyakarta, Java. There is also a version made with pumpkin. The geplak industry is mostly found in Bantul Regency, where it is commonly produced by home industries. Over time, this type of snack has spread widely due to market demand, and is now produced not only in the Special Region of Yogyakarta but also throughout the Indonesian archipelago.

\hrulefill
\vspace{1mm}

Extracted factual statements:
\vspace{1mm}

Geplak is a traditional Javanese snack from Bantul Regency, Special Region of Yogyakarta. The geplak industry is mostly run by home-based businesses in Bantul and has since spread throughout the Indonesian archipelago.

\end{tcolorbox}

\newpage
\section{Prompt for IndoCulture Benchmark}
\label{sec:prompt_benchmark}

The prompt used is the Indonesian MCQ prompt with province name as the location context, taken from the IndoCulture paper \citep{kotoIndoCultureExploringGeographically2024}.

\begin{tcolorbox}[title=IndoCulture MCQ Prompt]
\small
Untuk konteks [PROVINCE], sambungan yang tepat dari kalimat "[PREMISE]" adalah 
\vspace{1mm}

[OPTIONS]
\vspace{1mm}

Jawaban:

\hrulefill
\vspace{1mm}

English translation:
\vspace{1mm}

Given [PROVINCE] context, the correct continuation of the sentence "[PREMISE]" is
\vspace{1mm}

[OPTIONS]
\vspace{1mm}

Answer:
\end{tcolorbox}

\section{RAG-related prompts}
\label{sec:prompt_rag}

The prompt used for our RAG experiments is as follows:

\begin{tcolorbox}[title=Prompt for RAG]
\small
INSTRUKSI: Jawablah SOAL di bawah ini dengan bantuan BACAAN di bawah ini.
\vspace{1mm}

[DOCUMENT]
\vspace{1mm}

SOAL
\vspace{1mm}

[QUESTION]

\hrulefill
\vspace{1mm}

English translation:
\vspace{1mm}

INSTRUCTION: Answer the QUESTION below with the help of the PASSAGE below.

[DOCUMENT]
\vspace{1mm}

QUESTION
\vspace{1mm}

[QUESTION]

\end{tcolorbox}

The [QUESTION] field is replaced with an IndoCulture MCQ prompt given in Appendix \ref{sec:prompt_benchmark}.
The [DOCUMENT] field is replaced by the passages that are retrieved from the external corpus. Each passage added to the prompt is formatted as follows:

\begin{tcolorbox}
\small
BACAAN [DOC\_NUM]:
\vspace{1mm}

[DOC\_TEXT]
\end{tcolorbox}

The following prompt is used to generate the hypothetical document that may answer a question from IndoCulture. The [QUESTION] field is replaced with an IndoCulture MCQ prompt.  

\begin{tcolorbox}[title=Prompt for Hypothetical Document Generation]
\small
Write a passage in Indonesian language to answer the following question in detail. 
\vspace{1mm}

QUESTION:
\vspace{1mm}

[QUESTION]
\vspace{1mm}

PASSAGE:
\end{tcolorbox}

\section{Ontology of Journal Topics from Directory of Open Access Journals}
\label{sec:ontology_doaj}
\begin{itemize}
    \item \textbf{Fine Arts}
    \item \textbf{Geography. Anthropology. Recreation}
    \begin{itemize}
        \item Anthropology
        \item Environmental sciences
        \item Geography (General)
        \item Recreation. Leisure
        \begin{itemize}
            \item Dancing
        \end{itemize}
    \end{itemize}
    \item \textbf{Language and Literature}
    \begin{itemize}
        \item Literature (General)
        \item Philology, Linguistics
        \begin{itemize}
            \item Language, Linguistic theory, Comparative grammar
        \end{itemize}
    \end{itemize}
    \item \textbf{Music and Books on Music}
    \item \textbf{Philosophy. Psychology. Religion}
    \begin{itemize}
        \item Ethics
        \item Religions. Mythology. Rationalism     \end{itemize}
    \item \textbf{Social Sciences}
    \begin{itemize}
        \item Communities. Classes. Races
        \item Social history and conditions. Social problems. Social reform
        \item Social pathology. Social and public welfare. Criminology
        \item Social sciences (General)
        \item Social sciences and state - Asia (Asian studies only)
        \item Sociology (General)
        \item The family. Marriage. Woman
    \end{itemize}
\end{itemize}

\newpage
\section{Model Sources}
\label{sec:model_details}

\begin{table}[ht]
\small
\begin{tabular}{@{}>{\raggedright\arraybackslash}p{0.5\linewidth}>{\raggedright\arraybackslash}p{0.5\linewidth}@{}}
\toprule
\multicolumn{1}{c}{Model} & \multicolumn{1}{c}{Source}          \\ 
\midrule
SEALLMs-v3-7B             & SeaLLMs/SeaLLMs-v3-7B               \\
SEALLMs-v3-7B-Chat& SeaLLMs/SeaLLMs-v3-7B-Chat          \\
Sailor2-L-8B              & sail/Sailor2-L-8B                   \\
Sailor2-L-8B-Chat         & sail/Sailor2-L-8B-Chat              \\
Sailor2-L-20B             & sail/Sailor2-L-20B                  \\
Sailor2-L-20B-Chat        & sail/Sailor2-L-20B-Chat             \\
Qwen-SEA\_LION-v4-32B-IT  & aisingapore/Qwen-SEA-LION-v4-32B-IT \\ 
\bottomrule
\end{tabular}
\caption{HuggingFace sources of the models tested in this study. }
\label{tab:model_sources}
\end{table}

\section{Hardware and Time Details}
\label{sec:hardware}
Our experiments were conducted using Nvidia A100 GPUs. We used up to four GPUs for one evaluation run on IndoCulture. The time taken for one evaluation run depends on the model size and the number of documents retrieved for RAG. The time taken ranges from under one minute with a 7B model and no RAG, to around seven hours with a 32B models and 20 retrieved documents per question.

\onecolumn
\section{List of Journals in the Dataset}
\label{sec:all_crawled_journals}

\begin{longtable}{@{}ll@{}}
\toprule
Number & Journal Name                                                            \\
\midrule
1      & ANDHARUPA Jurnal Desain Komunikasi Visual \& Multimedia                 \\
2      & ARISTO                                                                  \\
3      & ARSNET                                                                  \\
4      & AT-TURAS Jurnal Studi Keislaman                                         \\
5      & Abdihaz Jurnal Ilmiah Pengabdian pada Masyarakat                        \\
6      & Absorbent Mind                                                          \\
7      & Academic Journal of Psychology and Counseling                           \\
8      & Al-Mazaahib Jurnal Perbandingan Hukum                                   \\
9      & Al-Misykah Jurnal Studi Al-qur'an dan Tafsir                            \\
10     & Analitika Jurnal Magister Psikologi UMA                                 \\
11     & Anthropos Jurnal Antropologi Sosial dan Budaya                          \\
12     & Arsitekno                                                               \\
13     & Arsitektura Jurnal Ilmiah Arsitektur dan Lingkungan Binaan              \\
14     & Az-Zahra Journal of Gender and Family Studies                           \\
15     & Basastra                                                                \\
16     & Biokultur                                                               \\
17     & Brikolase Jurnal Kajian Teori, Praktik dan Wacana Seni Budaya Rupa      \\
18     & Buddayah Jurnal Pendidikan Antropologi                                  \\
19     & Buletin Psikologi                                                       \\
20     & Buletin Riset Psikologi dan Kesehatan Mental (BRPKM)                    \\
21     & Bulletin of Counseling and Psychotherapy                                \\
22     & CaLLs (Journal of Culture, Arts, Literature, and Linguistics)           \\
23     & Dewa Ruci Jurnal Pengkajian dan Penciptaan Seni                         \\
24     & Dinamisia Jurnal Pengabdian Kepada Masyarakat                           \\
25     & EL-FIKR Jurnal Aqidah dan Filsafat Islam                                \\
26     & ENLIGHTEN Jurnal Bimbingan Konseling Islam                              \\
27     & ETHOS Jurnal Penelitian dan Pengabdian kepada Masyarakat                \\
28     & Edudeena Journal of Islamic Religious Education                         \\
29     & El-Aqwal Journal of Sharia and Comparative Law                          \\
30     & Engagement Jurnal Pengabdian Kepada Masyarakat                          \\
31     & GEMA TEOLOGIKA Jurnal Teologi Kontekstual dan Filsafat Keilahian        \\
32     & GUIDENA Jurnal Ilmu Pendidikan, Psikologi, Bimbingan dan Konseling      \\
33     & Gadjah Mada Journal of Professional Psychology (GamaJPP)                \\
34     & Gadjah Mada Journal of Psychology (GamaJoP)                             \\
35     & Gondang Jurnal Seni dan Budaya                                          \\
36     & Hanifiya Jurnal Studi Agama-Agama                                       \\
37     & Happiness Journal of Psychology and Islamic Science                     \\
38     & Harmoni Sosial Jurnal Pendidikan IPS                                    \\
39     & Hayula Indonesian Journal of Multidisciplinary Islamic Studies          \\
40     & Hisbah Jurnal Bimbingan Konseling dan Dakwah Islam                      \\
41     & Home Dynamics of Rural Society Journal                                  \\
42     & ICODEV Indonesian Community Development Journal                         \\
43     & INFERENSI Jurnal Penelitian Sosial Keagamaan                            \\
44     & INKLUSI                                                                 \\
45     & INSIGHT Jurnal Bimbingan Konseling                                      \\
46     & Ijtimā iyya Journal of Muslim Society Research                          \\
47     & Imajinasi Jurnal Seni                                                   \\
48     & Indonesian Journal of Earth Sciences                                    \\
49     & Indonesian Journal of Fundamental Sciences                              \\
50     & Indonesian Journal of Religion and Society                              \\
51     & Insight Jurnal Ilmiah Psikologi                                         \\
52     & International Journal Ihya' 'Ulum al-Din                                \\
53     & International Journal Pedagogy of Social Studies                        \\
54     & Islamic Counseling Jurnal Bimbingan Konseling Islam                     \\
55     & JADECS (Journal of Art, Design, Art Education \& Cultural Studies)      \\
56     & JAMBURA GEO EDUCATION JOURNAL                                           \\
57     & JAUR (JOURNAL OF ARCHITECTURE AND URBANISM RESEARCH)                    \\
58     & JIP (Jurnal Intervensi Psikologi)                                       \\
59     & JOINS (Journal of Information System)                                   \\
60     & JSW (Jurnal Sosiologi Walisongo)                                        \\
61     & JURNAL GEOGRAFI                                                         \\
62     & JURNAL PENELITIAN PENDIDIKAN, PSIKOLOGI DAN KESEHATAN (J-P3K)           \\
63     & JURNAL SOSIAL HUMANIORA (JSH)                                           \\
64     & Journal An-Nafs Kajian Penelitian Psikologi                             \\
65     & Journal Fenomena                                                        \\
66     & Journal Sampurasun                                                      \\
67     & Journal of Community Service and Empowerment                            \\
68     & Journal of Comparative Study of Religions                               \\
69     & Journal of Indonesian Society Empowerment                               \\
70     & Journal of Islamic Accounting and Finance Research                      \\
71     & Jurnal Adabiyah                                                         \\
72     & Jurnal Antropologi Isu-Isu Sosial Budaya                                \\
73     & Jurnal Dakwah Risalah                                                   \\
74     & Jurnal Diversita                                                        \\
75     & Jurnal EDUCATIO Jurnal Pendidikan Indonesia                             \\
76     & Jurnal Ekologi, Masyarakat dan Sains                                    \\
77     & Jurnal Humanitas Katalisator                                            \\
78     & Jurnal IPTA (Industri Perjalanan Wisata)                                \\
79     & Jurnal Ilmiah Pendidikan Pancasila dan Kewarganegaraan                  \\
80     & Jurnal Ilmiah Platax                                                    \\
81     & Jurnal Kajian Seni                                                      \\
82     & Jurnal Kawistara                                                        \\
83     & Jurnal Layanan Masyarakat (Journal of Public Services)                  \\
84     & Jurnal Litbang Provinsi Jawa Tengah                                     \\
85     & Jurnal Manusia dan Lingkungan                                           \\
86     & Jurnal Master Pariwisata (JUMPA)                                        \\
87     & Jurnal Pariwisata                                                       \\
88     & Jurnal Pariwisata Pesona                                                \\
89     & Jurnal Pariwisata Terapan                                               \\
90     & Jurnal Pembangunan Wilayah dan Kota                                     \\
91     & Jurnal Pemberdayaan Masyarakat Madani (JPMM)                            \\
92     & Jurnal Pemberdayaan Masyarakat Media Pemikiran dan Dakwah Pembangunan   \\
93     & Jurnal Psikoedukasi dan Konseling                                       \\
94     & Jurnal Psikogenesis                                                     \\
95     & Jurnal Psikologi Integratif                                             \\
96     & Jurnal Psikologi Islam dan Budaya                                       \\
97     & Jurnal Psikologi Teori dan Terapan                                      \\
98     & Jurnal Psikologi Ulayat                                                 \\
99     & Jurnal Riptek                                                           \\
100    & Jurnal Sains Psikologi                                                  \\
101    & Jurnal Sosiologi Andalas                                                \\
102    & Jurnal Sosiologi Pendidikan Humanis                                     \\
103    & Jurnal Sosiologi Reflektif                                              \\
104    & Jurnal Studi Agama                                                      \\
105    & KAIBON ABHINAYA JURNAL PENGABDIAN MASYARAKAT                            \\
106    & KLITIKA Jurnal Ilmiah Pendidikan Bahasa dan Sastra Indonesia            \\
107    & Kanz Philosophia A Journal for Islamic Philosophy and Mysticism         \\
108    & Khazanah Jurnal Studi Islam dan Humaniora                               \\
109    & Kifah Jurnal Pengabdian Masyarakat                                      \\
110    & LINGUA Jurnal Bahasa, Sastra, dan Pengajarannya                         \\
111    & Lamahu Jurnal Pengabdian Masyarakat Terintegrasi                        \\
112    & Linguistika                                                             \\
113    & MOZAIK HUMANIORA                                                        \\
114    & MUHARRIK Jurnal Dakwah dan Sosial                                       \\
115    & Majalah Geografi Indonesia                                              \\
116    & Masyarakat, Kebudayaan dan Politik                                      \\
117    & Moderatio Jurnal Moderasi Beragama                                      \\
118    & Mudra Jurnal Seni Budaya                                                \\
119    & Musãwa Jurnal Studi Gender dan Islam                                    \\
120    & NALARs                                                                  \\
121    & Nurani jurnal kajian syari'ah dan masyarakat                            \\
122    & POPULIKA                                                                \\
123    & PROMUSIKA                                                               \\
124    & Patra Widya Seri Penerbitan Penelitian Sejarah dan Budaya               \\
125    & Pelataran Seni                                                          \\
126    & Populasi                                                                \\
127    & Psikis Jurnal Psikologi Islami                                          \\
128    & Psikodimensia Kajian Ilmiah Psikologi                                   \\
129    & Psikoislamedia Jurnal Psikologi                                         \\
130    & Psikologika Jurnal Pemikiran dan Penelitian Psikologi                   \\
131    & Psympathic Jurnal Ilmiah Psikologi                                      \\
132    & QALAMUNA Jurnal Pendidikan, Sosial, dan Agama                           \\
133    & RUANG Jurnal Lingkungan Binaan (SPACE Journal of the Built Environment) \\
134    & Religi Jurnal Studi Agama-agama                                         \\
135    & Religious Jurnal Studi Agama-Agama dan Lintas Budaya                    \\
136    & Resital Jurnal Seni Pertunjukan                                         \\
137    & Riau Journal of Empowerment                                             \\
138    & SINTHOP Media Kajian Pendidikan, Agama, Sosial dan Budaya               \\
139    & Sawwa Jurnal Studi Gender                                               \\
140    & Simulacra                                                               \\
141    & Societas Dei Jurnal Agama dan Masyarakat                                \\
142    & SocioEdu Sociological Education                                         \\
143    & Soshum Jurnal Sosial dan Humaniora                                      \\
144    & Sosial Budaya                                                           \\
145    & Sosio-Didaktika Social Science Education Journal                        \\
146    & Tazkiya Journal of Psychology                                           \\
147    & VISIO DEI JURNAL TEOLOGI KRISTEN                                        \\
148    & Warta LPM                                                               \\
149    & Wawasan Jurnal Ilmiah Agama dan Sosial Budaya                           \\
150    & Zuriah Jurnal Pendidikan Anak Usia Dini                                 \\
151    & el Harakah Jurnal Budaya Islam                                          \\
\bottomrule
\end{longtable}

\end{document}